\documentclass{article} 
\usepackage{iclr2025_conference,times}

\usepackage{amsmath,amsfonts,bm}

\def\eqref#1{equation~\ref{#1}}

\def\1{\bm{1}}

\def\rvf{{\mathbf{f}}}
\def\rvg{{\mathbf{g}}}

\def\rvv{{\mathbf{v}}}

\def\rvx{{\mathbf{x}}}

\def\mI{{\bm{I}}}

\def\mM{{\bm{M}}}

\def\mR{{\bm{R}}}

\DeclareMathAlphabet{\mathsfit}{\encodingdefault}{\sfdefault}{m}{sl}
\SetMathAlphabet{\mathsfit}{bold}{\encodingdefault}{\sfdefault}{bx}{n}

\newcommand{\pdata}{p_{\rm{data}}}

\definecolor{niceblue}{rgb}{0.1,0.2,0.6}
\usepackage[pagebackref=true,breaklinks=true,colorlinks,bookmarks=true,linkcolor=niceblue,citecolor=niceblue,urlcolor=niceblue]{hyperref}

\usepackage{url}            
\usepackage{booktabs}       
\usepackage{amsfonts}       
\usepackage{nicefrac}       
\usepackage{microtype}      
\usepackage{xcolor}         
\usepackage{colortbl}
\usepackage{graphicx}
\usepackage{amsmath,amsfonts,amssymb,amsthm}
\usepackage{color}
\usepackage{caption}
\usepackage{subcaption}
\usepackage{xspace}
\usepackage{array}
\usepackage{tikzducks}
\usepackage{wrapfig}

\newcommand\sref{\S\ref}

\newcommand\eref{Eq.~\ref}
\newcommand\fref{Fig.~\ref}
\newcommand\tref{Table~\ref}
\usepackage{hyperref}       
\usepackage{url}            
\usepackage{booktabs}       
\usepackage{amsfonts}       
\usepackage{nicefrac}       
\usepackage{microtype}      
\usepackage{xcolor}         
\usepackage{float}
\usepackage{bm}
\usepackage{amsmath}
\usepackage{xspace}
\usepackage{graphicx}
\usepackage{enumitem}
\usepackage{subcaption}  
\usepackage{multirow}
\usepackage{caption}
\usepackage{threeparttable}
\usepackage{tablefootnote}
\usepackage{eucal}
\DeclareMathOperator{\rmsd}{RMSD}

\newcommand{\pref}{p_{\mathrm{ref}}}

\newcommand{\statespace}{S}

\newcommand{\kdelta}[2]{\delta \left\{ #1, #2 \right\} }

\newcommand{\noisemarg}{p_{t|1}}

\title{Applications of Modular Co-Design  \\ for De Novo 3D Molecule Generation}

\author{Danny Reidenbach$^\eta$\thanks{Equal Contributions} \hskip1em Filipp Nikitin$^\beta{}^\chi{}^*$\thanks{Work performed during internship at NVIDIA} \hskip1em  Olexandr Isayev$^\beta{}^\chi$ \hskip1em Saee Paliwal$^\eta$\\
  $^\eta$NVIDIA   \hskip1em  
  $^\beta$Department of Computational Biology, Carnegie Mellon University   \\
  $^\chi$Department of Chemistry, Carnegie Mellon University   \\
  \texttt{dreidenbach@nvidia.com} \\
}

\newcommand{\model}{\textnormal{Megalodon }}
\newcommand{\modelns}{\textnormal{Megalodon}}
\def\ie{{\it i.e.}}

\usepackage[toc,page,header]{appendix}
\usepackage{minitoc}

\usepackage{float}

\iclrfinalcopy 
\begin{document}

\maketitle

\doparttoc 
\faketableofcontents 

\begin{abstract}
De novo 3D molecule generation is a pivotal task in drug discovery. However, many recent geometric generative models struggle to produce high-quality 3D structures, even if they maintain 2D validity and topological stability. 
To tackle this issue and enhance the learning of effective molecular generation dynamics, we present \modelns--a family of scalable transformer models.
These models are enhanced with basic equivariant layers and trained using a joint continuous and discrete denoising co-design objective.
We assess \modelns's performance on established molecule generation benchmarks and introduce new 3D structure benchmarks that evaluate a model's capability to generate realistic molecular structures, particularly focusing on energetics.
We show that \model achieves state-of-the-art results in 3D molecule generation, conditional structure generation, and structure energy benchmarks using diffusion and flow matching. 
Furthermore, doubling the number of parameters in Megalodon to 40M significantly enhances its performance, generating up to 49x more valid large molecules and achieving energy levels that are 2-10x lower than those of the best prior generative models.
\end{abstract}
\section{Introduction}
Molecular Generative models have been heavily explored due to the allure of enabling efficient virtual screening and targeted drug design~\citep{gomez-vae}. Similar to the rise in their application to computer vision (CV)~\citep{DiT, sit}, Diffusion and Flow Matching models have been applied for tasks including molecule design, molecular docking, and protein folding~\citep{diffsbdd, diffdock, af3}. Across CV and chemical design, the scaling of model architectures and training data have seen significant accuracy improvements but questions surrounding how to scale effectively still persist~\citep{plinder}.

Specifically for 3D molecule generation (3DMG), where the task is to unconditionally generate valid and diverse 3D molecules, diffusion models have shown great promise in enabling accurate generation starting from pure noise~\citep{EDM}. The iterative nature of diffusion models allows them to explore a diverse range of molecular configurations, ideally providing valuable insights into potential drug candidates and facilitating the discovery of novel compounds. However, unlike in CV, which has seen systematic evaluations of training data and scaling, with tangible benchmark results~\citep{esser2024scaling}, measuring success in de novo molecule generation is quite difficult. As a result, there is a nonlinear path to determining what truly is making an impact if, in each model, the data, architecture, training objective, and benchmarks differ. Furthermore, the commonly shared 3DMG benchmarks that do exist only evaluate 2D quantities, ignoring 3D structure, conformational energy, and model generalization to large molecule sizes--all quantities that are imperative for real-world use. In this work, we explore the above in the context of 3DMG and its interpretable benchmarks to directly target larger molecules.


\begin{figure*}[t]
\centering
    \includegraphics[width=0.75\textwidth]{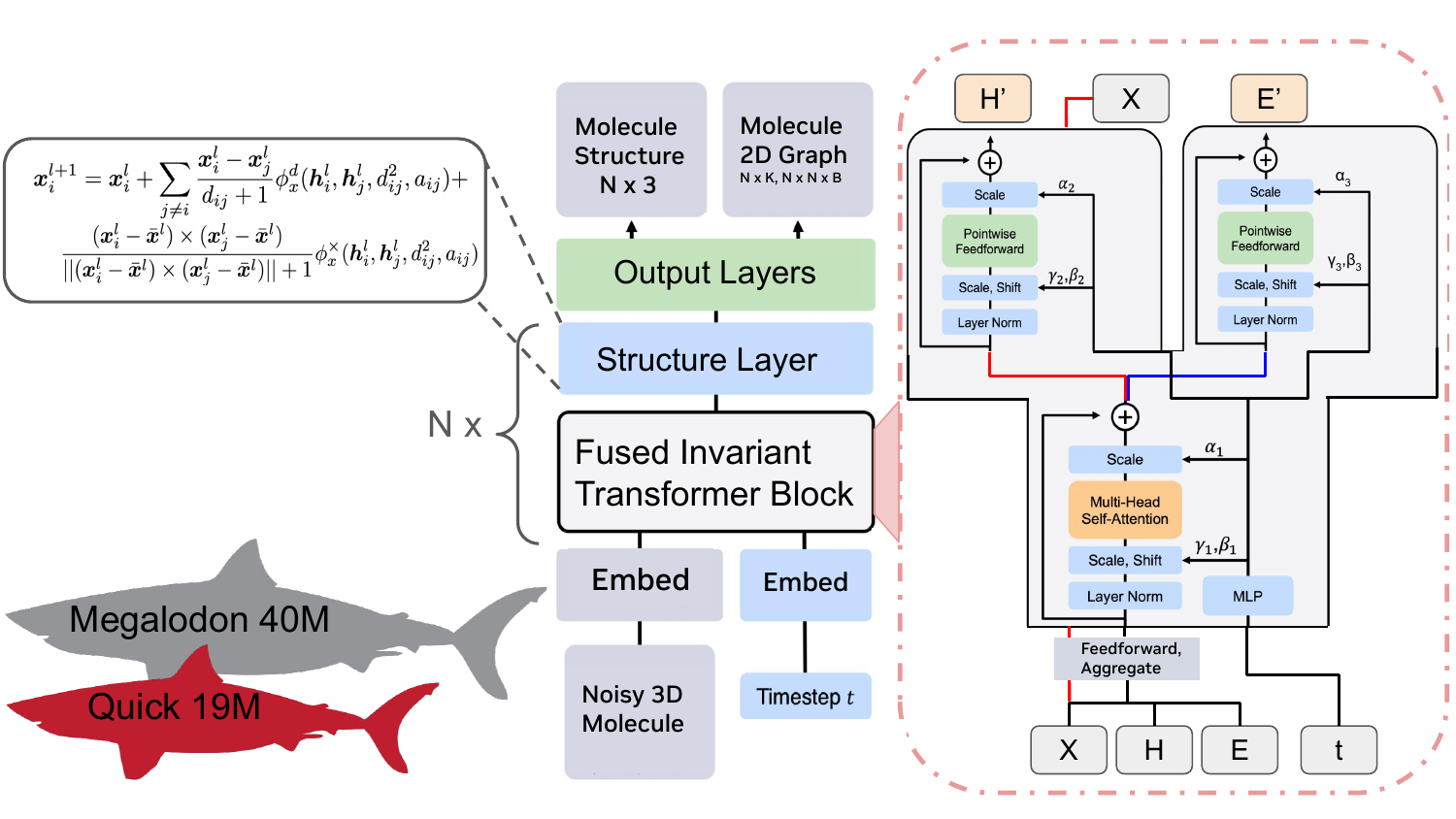}
    \caption{Megalodon Architecture: molecules are separated into 3D structures and discrete atom types, bond types, and atom charge features. All features are embedded separately, passed through a feed-forward layer, and aggregated to produce the input tokens for the fused Invariant Transformer blocks. The embedded structure features and transformer outputs for the discrete features are passed to a single EGNN layer for structure updates. The output heads consist of standard MLPs and an EGNN layer for bond refinement.}
    \label{fig:arch}
    \vspace{-3ex}
\end{figure*}

Our main contributions are as follows:
\vspace{-2ex}
\begin{itemize}
\item We present \modelns, a scalable transformer-based architecture for multi-modal molecule diffusion and flow matching. This is the first 3DMG model to be tested with both objectives, with both obtaining state-of-the-art results. We show that our diffusion model excels at structure and energy benchmarks, whereas our flow matching model yields better 2D stability and the ability to use 25x fewer inference steps than its diffusion counterpart.
\item \model is the first model capable of unconditional molecule generation and conditional structure generation without retraining or finetuning. 
\item We introduce new geometric benchmarks focusing on the interpretable physics-based quantum mechanical (QM) and molecular conformational energy.
\end{itemize}
\vspace{-2ex}

\vspace{-1ex}
\section{Background}
\label{sec:rw}
\subsection{3D Molecule Generation}
In de novo 3D molecule generation (3DMG), a molecule's 3D structure and 2D topology are simultaneously generated. We define a molecule $\mM = (X, H, E, C)$ with $N$ atoms where $X\in\mR^{N \times 3}$,  $H\in\{0,1\}^{N \times A}$, $E\in\{0,1\}^{N \times N \times B}$, and $C\in\{0,1\}^{N \times K}$ represents the atom coordinates, element types, bond types adjacency matrix, and formal charges respectively. $X$ is modeled as a continuous variable whereas $H$, $E$, and $C$ are discrete one-hot variables. 
\vspace{-1ex}
\subsection{Important Qualities of 3D Molecules}
The GEOM dataset~\citep{geom} is widely used for 3D molecular structure (conformer) generation tasks, containing 3D conformations from both the QM9 and drug-like molecule (DRUGS) databases, with the latter presenting more complex and realistic molecular challenges. Conformers in the dataset were generated using CREST~\citep{crest}, which performs extensive conformational sampling based on the semi-empirical extended tight-binding method (GFN2-xTB)~\citep{xtb}. This ensures that each conformation represents a local minimum in the GFN2-xTB energy landscape.

Energy, in the context of molecular conformations, refers to the potential energy of a molecule's structure, which is a key determinant of its stability. Lower-energy conformations are typically more stable and are found at the minima on the potential energy surface (PES). For a generative model to succeed, it must not only generate molecules that are chemically valid but also ones that correspond to low-energy conformations, reflecting local minima on the PES. Thus, energy serves as the ultimate measure of success in molecular modeling, as it directly correlates with the physical realism and stability of the generated structures.

A key requirement for generative models is their ability to implicitly learn this energy landscape and produce molecules that are local minima of the potential energy surface. However, since GFN2-xTB is itself a model rather than a universal energy function, comparing energies across different potentials (e.g., using GFN2-xTB optimized structures but computing energies with MMFF~\citep{mmff}) can introduce systematic errors. Differences in potential models, such as optimal bond lengths, may lead to unreliable results.
Overall, the goal of 3DMG is to generate valid and low-energy molecules mimicking the energy landscape of the GEOM dataset.

\vspace{-1ex}
\subsection{Related Work}
\citet{EDM} first introduced continuous diffusion modeling for coordinates and atom types using a standard EGNN architecture~\citep{egnn}. Following this, many models have been produced that make slight changes to the architecture and interpolant schedule to generate atom coordinates and types~\citep{EquiFM}. While initially effective, they rely on the molecule-building software OpenBabel~\citep{openbable} to infer and update the bond locations and types, which is a critical aspect of the stability calculations. The issues and biases of OpenBabel have been heavily explored, and as a result, methods began to generate the bond locations and types in the generative process~\citep{pat_walters_blog}. \citet{MiDi} was the first to use continuous diffusion for coordinates and discrete diffusion for the atom and bond types, removing the OpenBabel requirement. \citet{eqgat} used the same training objective but introduced a more effective equivariant architecture. Recently \citet{semla} uses continuous and discrete flow matching with a latent equivariant graph message passing architecture to show improved performance. For further discussion please see~\sref{sec:app-rw}

\vspace{-1ex}
\subsection{Stochastic Interpolants}
\paragraph{Continuous Gaussian Interpolation}
Following~\citep{lipman2023flow,albergo2023stochastic}, in the generative modeling setting, we construct interpolated states between an empirical data and a Gaussian noise distribution $\mathcal{N}(\rvx_t; \beta(t)x_1, \alpha(t)^2\mI)
$, this is,
\begin{subequations} \label{eq:interpolant}
\begin{align}
    \rvx_t &= \alpha(t)\boldsymbol{\epsilon} + \beta(t)\rvx_1, \label{eq:interpolant_a} \\
    \rvx_1 &= \frac{\rvx_t-\alpha(t)\boldsymbol{\epsilon}}{\beta(t)} \label{eq:interpolant_b}
\end{align}
\end{subequations}

where $\boldsymbol{\epsilon}\sim\mathcal{N}(\boldsymbol{\epsilon};\mathbf{0},\mI)$ and $\rvx_1\sim p_\textrm{data}(\rvx_1)$.
Common choices for the interpolation include (assuming $t\in[0,1]$), with $t=1$ corresponding to data and $t=0$ to noise:
\vspace{-2ex}
\begin{itemize}
    \item Variance-preserving SDE-like from the diffusion model literature~\citep{song2020score}: $\alpha(t) = \sqrt{1-\gamma_t^2}$ and $\beta(t) = \sqrt{\gamma_t^2}$ with some specific ``noise schedule'' $\gamma_t$ which is commonly written as $\sqrt{\Bar{\alpha_t}}$ from \citet{DDPM}.
        \item Conditional linear vector field~\citep{lipman2023flow}: $\alpha(t)=1-(1-\sigma_{\textrm{min}})t$ and $\beta(t)=t$ with some smoothening of the data distribution $\sigma_{\textrm{min}}$. 
\end{itemize}

\vspace{-2ex}
\paragraph{Continuous Diffusion}
Continuous Denoising Diffusion Probabilistic Models (DDPM) integrate a gradient-free forward noising process based on a predefined discrete-time variance schedule (~\eref{eq:interpolant_a}) and a gradient-based reverse or denoising process~\citep{DDPM}. The denoising model can be parameterized by data or noise prediction as they can be equilibrated via~\eref{eq:interpolant_b}. Following \citet{eqgat},  we use the following training objective and update rule:
\begin{equation}
\begin{split}
    \mathcal{L}_\textrm{DDPM}(\theta)&=\mathbb{E}_{t,\boldsymbol{\epsilon}\sim\mathcal{N}(\boldsymbol{\epsilon};\mathbf{0},\mI),\rvx_1\sim p_\textrm{data}(\rvx_1)}||\rvx_\theta(t,\rvx_t)-\rvx_1||^2 \\
\end{split}
\end{equation}
\begin{equation}
\small
\begin{split}
    \mu_\theta(t,\rvx_t)&= \rvf(\alpha(t), \beta(t))*\rvx_\theta(t,\rvx_t) + \rvg(\alpha(t), \beta(t))*\rvx_t \\
    \rvx_{t+1}&= \mu_\theta(t,\rvx_t) + \sigma((\alpha(t), \beta(t))*\epsilon 
\end{split}
\vspace{-2ex}
\end{equation}
where functions $\textbf{f}$, $\textbf{g}$, and $\boldsymbol{\sigma}$ are defined for any noise schedule such as the cosine noise schedule used in~\citet{MiDi}.
\vspace{-1ex}
\paragraph{Continuous Flow Matching}
Flow matching (FM) models are trained using the conditional flow matching (CFM) objective to learn a time-dependent vector field $\rvv_\theta(t,\rvx_t)$ derived from a simple ordinary differential equation (ODE) that pushes samples from an easy-to-obtain noise distribution to a complex data distribution.
\begin{equation}
\small
\begin{split}
    \mathcal{L}_\textrm{CFM}(\theta)&=\mathbb{E}_{t,\boldsymbol{\epsilon}\sim\mathcal{N}(\boldsymbol{\epsilon};\mathbf{0},\mI),\rvx_1\sim p_\textrm{data}(\rvx_1)}||\rvv_\theta(t,\rvx_t)-\frac{d}{dt}\rvx_t||^2 \\
    & =\mathbb{E}_{t,\boldsymbol{\epsilon}\sim\mathcal{N}(\boldsymbol{\epsilon};\mathbf{0},\mI),\rvx_1\sim p_\textrm{data}(\rvx_1)}||\rvv_\theta(t,\rvx_t)-\dot{\alpha}(t)\boldsymbol{\epsilon} - \dot{\beta}(t)\rvx_1||^2,
\end{split}
\end{equation}
The time-differentiable interpolation seen in \eref{eq:interpolant_a} gives rise to a probability path that can be easily sampled. For more details on how to relate the Gaussian diffusion and CFM objectives with the underlying score function of the data distribution, please see Appendix~\ref{sec:app-diff}.

In practice, many methods use a "data prediction" objective to simplify training, which gives rise to the following loss function and inference Euler ODE update step following the conditional linear vector field~\citep{lipman2023flow, semla}.
\begin{equation}
\small
\begin{split}
    \mathcal{L}_\textrm{CFM}(\theta)&=\mathbb{E}_{t,\boldsymbol{\epsilon}\sim\mathcal{N}(\boldsymbol{\epsilon};\mathbf{0},\mI),\rvx_1\sim p_\textrm{data}(\rvx_1)}||\rvx_\theta(t,\rvx_t)-\rvx_1||^2 \\
\end{split}
\end{equation}
\begin{equation}
\small
\begin{split}
    \rvv_\theta(t,\rvx_t)&= \frac{\rvx_\theta(t,\rvx_t) - x_t}{1-t}, \\
    \rvx_{t+1}&= x_t + \rvv_\theta(t,\rvx_t)dt
\end{split}
\end{equation}

\paragraph{Discrete Diffusion} Following~\citet{d3pm}, Discrete Denoising Diffusion Probabilistic Models (D3PMs) apply the same concept as continuous diffusion but over a discrete state space. Like the continuous counterpart that relies on a predefined schedule to move mass from the data to prior distribution, D3PM uses a predefined transition matrix that controls how the model transitions from one discrete state to another.

For scalar discrete random variables with $K$ categories $x_t, x_{t-1} \in {1, ..., K}$
the forward transition probabilities can be represented by matrices: $[{Q}_{t}]_{ij} = q(x_{t}=j|x_{t+1}=i)$.
Starting from our data $x_1$ or $x_T$ (where $T$ is the total number of discrete time steps)\footnote{We adjust the direction of time for diffusion to match the FM equations such that T=1 is data.}, we obtain the following $T-t+1$ step marginal and posterior at time $t$:
\begin{equation}
\small
\begin{split}
    &\overline{ Q}_{t} =  Q_t  Q_{t+1} \hdots  Q_T \\
    & q(x_{t}|x_{t+1}) = \mathrm{Cat}(x_{t}; p = x_{t+1} Q_{t}), \quad q(x_t | x_T) = \mathrm{Cat}\left(x_{t}; p = x_{T}\overline{ Q}_{t}  \right), \\
    & q(x_{t+1}|x_{t}, x_T) = \frac{q(x_{t}|x_{t+1}, x_T)q(x_{t+1}|x_T)}{q(x_{t}| x_T)} \\
    &=\mathrm{Cat}\left(x_{t+1}; p= \frac{x_{t} Q_{t}^{\top} \odot  x_T\overline{Q}_{t+1}  }{x_T \overline{Q}_{t} x_{t}^\top}\right)\label{eq:discrete_t_step_posterior}
\end{split}
\end{equation}
Here $Q$ is defined as a function of the same cosine noise schedule used in continuous DDPM such that the discrete distribution converges to the desired terminal distribution (\ie uniform prior) in T discrete steps. Similar to the use of mean squared error loss for DDPM, D3PM uses a discrete cross-entropy objective.
\vspace{-1ex}
\paragraph{Discrete Flow Matching} Following~\citet{multiflow}, we use the Discrete Flow Matching (DFM) framework to learn conditional flows for the discrete components of molecule generation ( atom types, bond types, and atom charges). We use the following DFM interpolation in continuous time, where $S$ is the size of the discrete state space:
\begin{equation}
    \begin{split}
        \noisemarg^{\mathrm{unif}}(x_t | x_1) &= q(x_t | x_1) =\mathrm{Cat}(t \kdelta{x_1}{x_t} + (1-t) \frac{1}{\statespace}),
    \end{split}
\end{equation}
Similar to discrete diffusion, we use the cross-entropy objective for training. Please see~\citet{multiflow} for sampling procedure details.

\begin{figure}[htbp]
    \centering
    \begin{subfigure}[b]{0.49\textwidth}
        \centering
        \includegraphics[width=\textwidth]{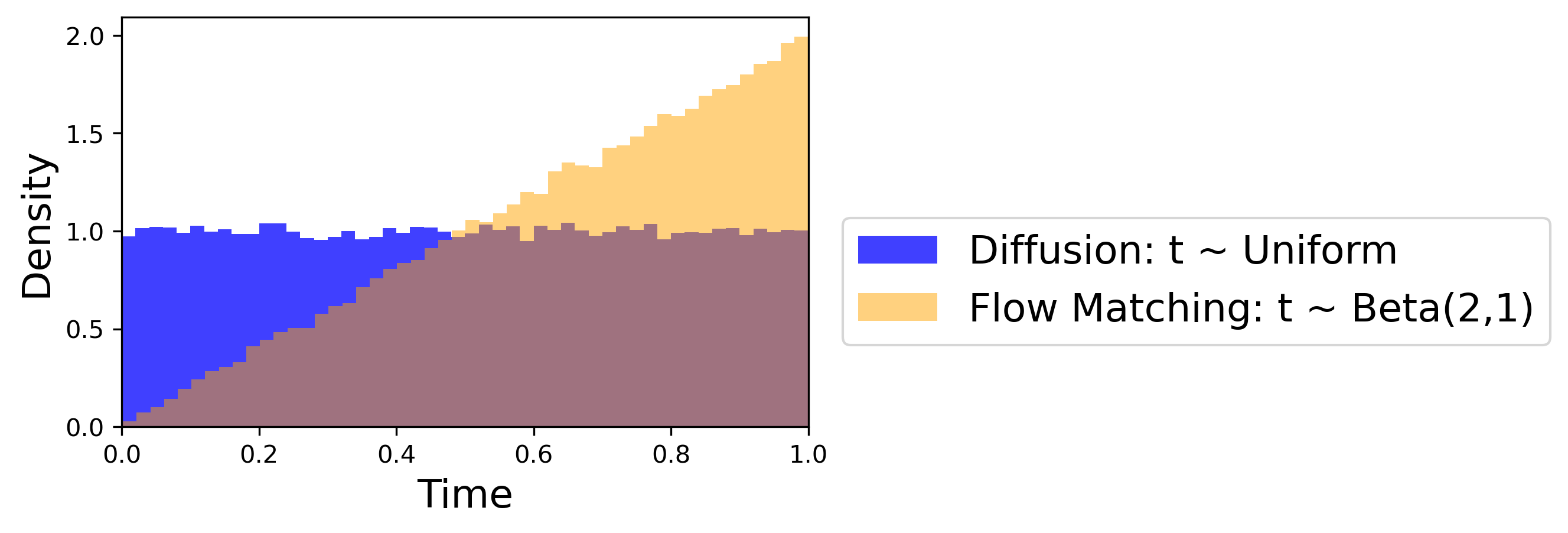}
        \caption{Time distributions used for molecule generation}
    \end{subfigure}
    \hfill
    \begin{subfigure}[b]{0.49\textwidth}
        \centering
        \includegraphics[width=\textwidth]{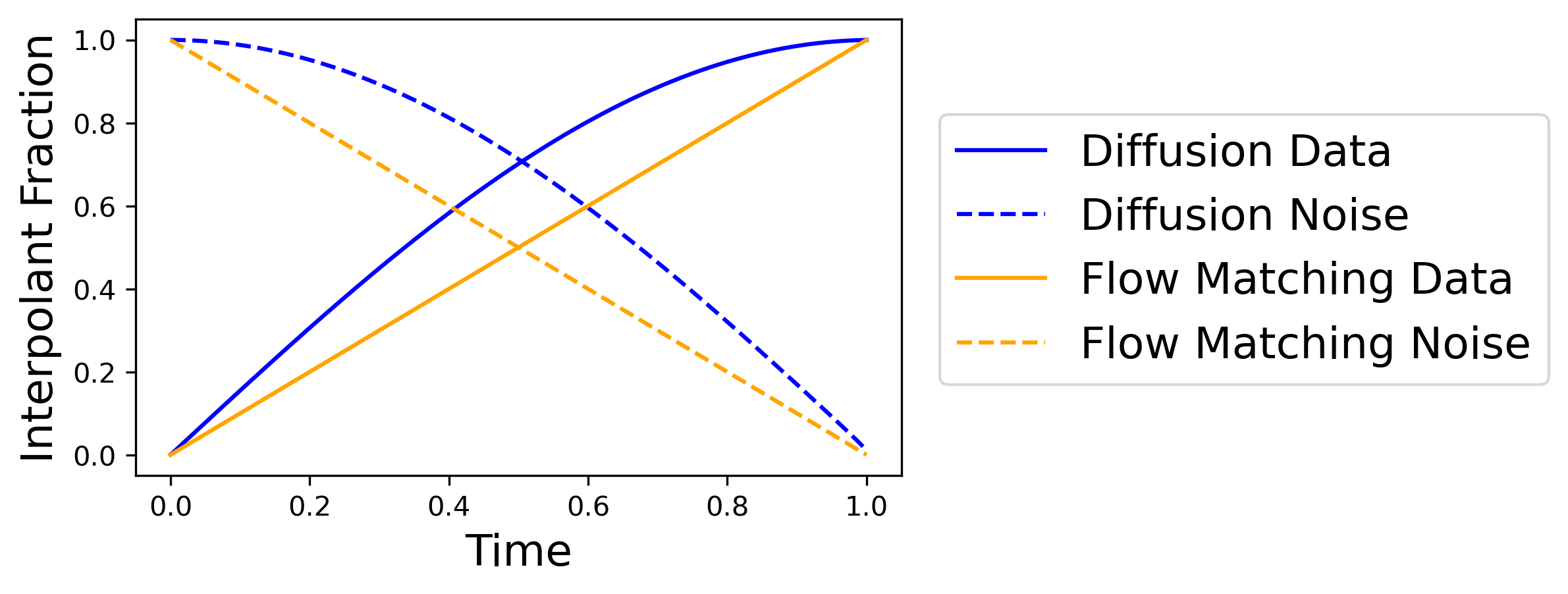}
        \caption{FM linear vs. Diffusion cosine interpolant}
    \end{subfigure}
    \caption{Time and interpolation comparison between \model and \modelns-flow}
    \label{fig:fmvdiff}
    \vspace{-1ex}
\end{figure}

\vspace{-1ex}
\paragraph{Diffusion vs. Flow Matching} 
We see that for both Diffusion and CFM, the loss functions used in practice are identical. Differences arise in how we build the interpolation, how we sample from these models, and their theoretical constraints.
Diffusion models rely on complex interpolation schedules that are tuned to heavily weight the data distribution using a uniform time distribution. In contrast, FM commonly uses a simple linear interpolation but can achieve that same data distribution weighting by sampling from more complex time distributions. The choices of time distributions and interpolation schedules can be chosen appropriately to make FM and Diffusion equivalent in the Gaussian setting (see Sec.~\ref{sec:app-diff}). We show in~\fref{fig:fmvdiff} the interpolation and time distribution differences that mimic the same weighting of $\pdata$ at T=1 that are currently used in recent 3DMG models~\citep{eqgat, semla}.

Diffusion models inherently rely on simulating Gaussian stochastic processes. In the forward process, data points are progressively noised, converging towards a Gaussian distribution. This process, derived from score-based generative models, aims to learn the score function (the gradient of the data distribution's log density) to reverse the diffusion process. The generative model effectively solves a Stochastic Differential Equation (SDE) that describes how data diffuses towards noise and how it can be denoised in reverse. The reverse process requires SDE simulation at every step, which involves sampling from a learned probabilistic model that estimates how to remove noise. This involves simulating random variables at each time step, making diffusion models highly dependent on repeated stochastic simulation. 

Flow Matching, on the other hand, learns a continuous vector field that deterministically "flows" one distribution to another. The model learns this flow by matching the velocity field that pushes samples from a source distribution to a target distribution. Once the vector field is learned, generating samples involves solving an ODE that defines a continuous and deterministic trajectory from the source to the target distribution. Unlike diffusion models, which require simulating a series of stochastic transitions (noising and denoising) over many steps, flow matching learns a single, continuous flow. Sampling involves solving an ODE (or, in some cases, a deterministic SDE with noise) to move from the base distribution to the target in a smooth, deterministic fashion.

For DDPM, the equations only hold for the Gaussian path with access to a well-formed score function. This is why techniques like mini-batch Optimal Transport (OT) can be applied to FM but not Diffusion to align $\pdata$ and $\pref$~\citep{tong2023improving}. In FM, the vector field is learned, which, in the absence of OT, can be derived as a function of the score function, but having access to the score function is not a requirement to sample deterministically (simulation-free).

\vspace{-1ex}
\section{Methods} \label{sec:methods}
\paragraph{Megalodon Architecture} 
Since 3DMG allows for the simultaneous generation of a discrete 2D molecular graph and its 3D structure, we intentionally designed our architecture with a core transformer trunk to better model discrete data~\citep{attention, gpt3}. 
\fref{fig:arch} illustrates the model architecture, which is comprised of N blocks made up of fused invariant transformer blocks and simple structure update layers, followed by linear layers for discrete data projection. 

In the fused invariant transformer block, the embedded structure, atom types, and bond types are fused and aggregated to create a single molecule feature. This is passed into a standard multi-head attention module with adaptive layernorm. The scaled output is then passed into separate adaptive layernorm feedforward blocks for the atom types and bond types. The transformer also produces an unchanged molecule structure via a residual connection to the input. The updated atom and bond types are then passed into a simple structure layer. The structure layer only updates the predicted structure via a standard distance-based EGNN update with a cross-product term~\citep{egnn, diffsbdd}. We emphasize that this cross-product term is critical for model performance. At a high level, the transformer block updates our discrete invariant data, and our equivariant layer updates our structure. \model uses scaling tricks such as query key prenorm and equivariant norm in our structure layer~\citep{esm3}. We also note \modelns, at 4x more parameters, is more memory efficient than EQGAT-diff~\citep{eqgat}, enabling 2x larger batch size while still having the quadratic dependency of fully connected edge features. For more details, please see Sec.~\ref{sec:arch-app}.

We introduce a generative scaling benchmark, and as we show, the performance of 3DMG models is correlated with the size of the generated molecules. We note that our large model is, in fact, not that large compared to recent biological models~\cite {esm, esm3} and can be further scaled beyond 40M params if further benchmarks are developed.
\vspace{-1ex}
\paragraph{Training Objective} We explore \model in the context of diffusion and flow matching. For our diffusion flavored model, following \citet{eqgat, MiDi} we use the same weighted cosine noise schedules, DDPM, and discrete D3PM objective. 
When using conditional flow matching, we apply the same training objective and hyperparameters as \citet{semla}, including equivariant optimal transport. 
In this way, for diffusion and flow matching, we train and evaluate our model in an \emph{identical way} including hyperparameters to prior models of same types.

In our experiments with EQGAT-diff, we found that the diffusion objective with data-like priors possesses an interesting but potentially harmful behavior. 
Although the noise sample from the data-prior and the true data sample have bonds, the model consistently generates no bonds for all time $\leq0.5$, which corresponds to an interpolation with $\leq70\%$ of the data as seen in \fref{fig:fmvdiff}(b). Therefore there is no useful information for the edge features in half the training and inference samples. As a result, only when the structure error is low, as the model starts with 70\% data in the interpolation, does the bond prediction accuracy jump to near-perfect accuracy. Thus, only when the structure is accurate was the 2D graph accurate, which is counterintuitive to the independent and simultaneous objective. In other words, the 2D graph does not inform the 3D structure as one would expect to happen, and we would want equal importance on the 2D topology and 3D structure.

To address this inefficiency, as the structure, atom type, and bond type prediction inform each other to improve molecule generation, we introduce a subtle change to the training procedure similar to \citet{multiflow}. Keeping each data type having its own independent noise schedule, we enable a concrete connection between the discrete and continuous data that it is modeling. 
Explicitly, rather than sampling a single time variable, we introduce a second noise variable to create $t_{continuous}$ and $t_{discrete}$, both sampled from the same time distribution.
Now discrete and continuous data are interpolated with their respective time variable \emph{and} maintain the independent weighted noise schedules. 
We note that the MiDi weighted cosine schedules were already adding different levels of noise for the same time value. Now, we take that one step further and allow the model to fill in the structure given the 2D graph and learn to handle more diverse data interpolations.

\vspace{-1ex}
\paragraph{Self Conditioning}
Following~\citet{selfconditioning}, we train \model with self-conditioning similar to prior biological generative models~\citep{FrameDiff, harmonicflow, semla}. We found that constructing self-conditioning as an outer model wrapper with a residual connection led to faster training convergence:
\vspace{-2ex}
\begin{equation}
\begin{split}
x_{\text{sc}} &= \text{model}(x_t) \\
x_t &= \text{MLP}([x_{\text{sc}}, x_t]) + x_t \\
x_{\text{pred}} &= \text{model}(x_t)
\end{split}
\end{equation}
Specifically for 3DMG, self-conditioning is applied independently to each molecule component $\mM = (X, H, E, C)$, where the structure component uses linear layers without bias and all discrete components operate over the raw logits rather than the one-hot predictions.

\vspace{-1ex}
\section{Experiments}
\label{sec:exp}
\paragraph{Data} GEOM Drugs is a dataset of drug-like molecules with an average size of around 44 atoms~\citep{geom}. Following~\citet{eqgat}, we use the same training splits as ~\citet{MiDi}. 
We emphasize that the traditional metrics are calculated by first sampling molecule sizes from the dataset(~\fref{fig:node-dist}) and then generating molecules with the sampled number of atoms, including explicit hydrogens. 
We show in Sec.~\ref{sec:3dmg} that this does not illustrate the full generative capacity, as in many real-world instances, people want to generate molecules with greater than 100 atoms~\citep{protac}.

\vspace{-1ex}
\subsection{Unconditional De Novo Generation}
\label{sec:3dmg}
\paragraph{Problem Setup} Following \citet{eqgat} we generate 5000 molecules (randomly sampling the number of atoms from the train distribution see~\fref{fig:node-dist}), and report (1) Atom Stability: the percentage of individual atoms that have the correct valency according to its electronic configuration that was predefined in a lookup table, (2) Molecule Stability: percentage of molecules in which all atoms are stable, (3) Connected Validity: fraction of molecules with a single connected component which can be sanitized with RDKit.
We also introduce two structural distributional metrics for the generated data: (4) bond angles and (5) dihedral angles, calculated as the weighted sum of the Wasserstein distance between the true and generated angle distributions, with weights based on the central atom type for bond angles and the central bond type for dihedral angles, respectively. We note this is a high-level metric and not reducible to a physically meaningful per-molecule error. To combat this, please see our new physical structure metrics in Sec.~\ref{sec:xtb}.

\vspace{-1ex}
\paragraph{Baselines} EQGAT-diff has 12.3M parameters and leverages continuous and discrete diffusion~\citep{eqgat}. SemlaFlow has 23.3M params\footnote{Checkpoint from public code has 2 sets of 23.2M params, one for the last gradient step and EMA weights} and is trained with conditional flow matching with equivariant optimal transport~\citep{semla}. We report two \model sizes, small (19M) and large (40.6M). We train with identical objectives and settings to both EQGAT-diff and SemlaFlow. We also compare to older diffusion models, including MiDi and EDM, as they introduce imperative techniques from which the more recent models are built.

\begin{table*}[t]
\caption{Measuring Unconditional Molecule Generation: 2D and 3D benchmarks. * Denotes taken from EQGAT-Diff.}
\label{tab:3dmg}
\centering
\begin{tabular}{l c | c | c c c | c c}
\toprule
& \multicolumn{1}{l|}{} & \multicolumn{1}{c|}{} & \multicolumn{3}{c|}{2D Topological ($\uparrow$)} & \multicolumn{2}{c}{3D Distributional ($\downarrow$)}  \\
Model &  & Steps & Atom Stab. & Mol Stab. & Validity  & Bond Angle & Dihedral \\ \midrule
\multicolumn{2}{l|}{EDM+OpenBabel*} & 1000 & 0.978 & 0.403 & 0.363  & -- & -- \\
\multicolumn{2}{l|}{MiDi*} & 500 & 0.997 & 0.897 & 0.705  & -- & --  \\
\multicolumn{2}{l|}{EQGAT-diff$^{x0}_{disc}$} & 500 & 0.998 & 0.935 & 0.830 & 0.858 & 2.860 \\

\multicolumn{2}{l|}{EGNN + cross product} & 500 & 0.982 & 0.713 & 0.223  & 14.778 & 17.003 \\
\multicolumn{2}{l|}{\modelns-quick} & 500& 0.998 & 0.961 & 0.900  & 0.689 & 2.383 \\
\multicolumn{2}{l|}{\modelns} & 500& \textbf{0.999} & \textbf{0.977} & \textbf{0.927}  & \textbf{0.461} & \textbf{1.231} \\
\midrule
\multicolumn{2}{l|}{SemlaFlow} & 100 & \textbf{0.998} & 0.979 & 0.920 & 1.274 & \textbf{1.934} \\
\multicolumn{2}{l|}{\modelns-flow} & 100 & 0.997 & \textbf{0.990} & \textbf{0.948}  & \textbf{0.976} & 2.085 \\
\bottomrule
\end{tabular}
\vspace{-1ex}
\end{table*}

\vspace{-1ex}
\paragraph{Analysis} Both the diffusion and flow matching versions of \model achieve state-of-the-art results. With the FM version obtaining better topological accuracy and the diffusion version seeing significantly improved structure accuracy. This experiment shows that the underlying augmented transformer is useful for the discrete and continuous data requirements of 3DMG, regardless of the interpolant and sampling methodology. We also see that the transformer part is crucial for \modelns's success as just using the EGNN with cross-product updates with standard edge and feature updates for the non-equivariant quantities performs quite poorly. We also note that all methods obtain 100\% \emph{uniqueness}, 88-90\% \emph{diversity}, and 99\% \emph{novelty} following~\citep{eqgat} definitions with no meaningful performance differences. For further model comparisons and model ablations surrounding reducing the number of inference steps please see Appendix. \tref{tab:3dmg-app}. Please see \sref{sec:qm9} to see how \model performs on another task/dataset to demonstrate the generalizability.

\begin{figure}[t]
\centering
    \includegraphics[width=0.9\textwidth]{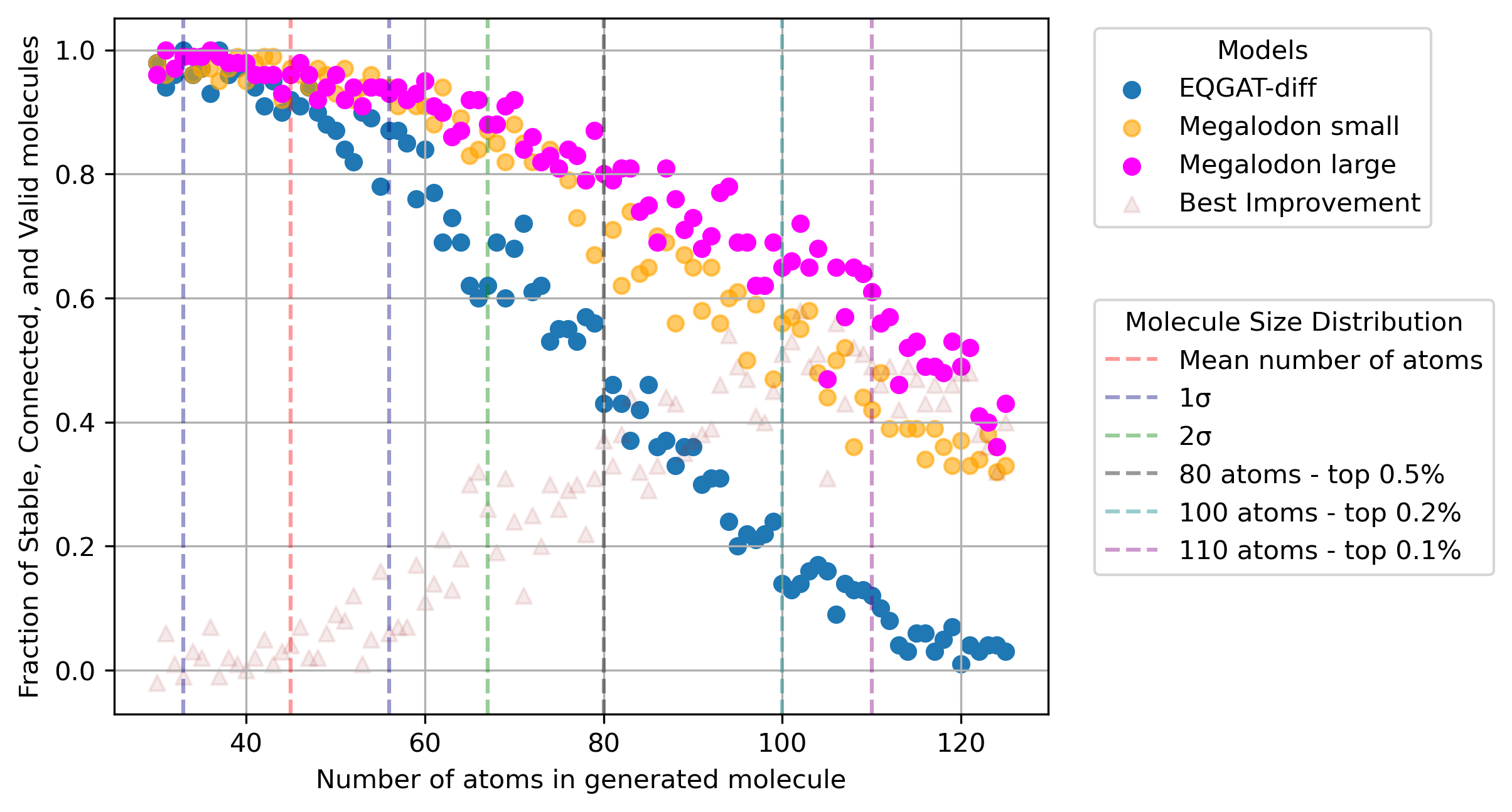}
    \caption{Diffusion model performance as a function of molecule size. Note the ability for \model to generate valid and stable molecules with little training data support.}
    \label{fig:val-of-len}
    \vspace{-5mm}
\end{figure}

\vspace{-1ex}
\paragraph{Impact of molecule size on performance}
As~\tref{tab:3dmg} shows average results over 5000 molecules of relatively small and similar sizes, it is hard to understand if the models are learning how to generate molecules or just regurgitating training-like data. We design an experiment to directly evaluate this question and see how models perform as they are tasked to generate molecules outside the support region of the train set.
We see in \fref{fig:val-of-len} that the topological model performance is a function of length (for full-size distribution, see~\fref{fig:node-dist}). Here for each length [30, 125] we generate 100 molecules and report the percentage of stable and connected valid molecules. 

We emphasize that \tref{tab:3dmg} illustrates only a slice of the performance via the average of 5K molecules sampled from the train set size distribution. We note that although molecules with greater than 72 atoms make up $\leq$ 1\% of the train set, \model demonstrates roughly 2-49x better performance than EQGAT-diff for the larger half of the generated molecule sizes. 
We hypothesize that since molecule stability is a discrete 2D measurement, the transformer blocks in \model allow it to better generalize even if seeing similar molecules in less than 0.1\% of the training data. 
In other words, the ability of transformers to excel at modeling discrete sequential data improves our generative performance.
We want to point out that all tested models are trained with identical datasets, hyperparameters, diffusion schedules, and training objectives. 
The only difference is the architecture.
We also see that the ability to scale our simple architecture allows the model to even better generate molecules outside the region of data support.
Lastly, we chose to focus on only the diffusion models here as they exhibit the best structure benchmark performance.

\begin{table}[!htbp]
    \centering
    \caption{Normalized Model Inference Speed}
    \begin{tabular}{l|c|c}
        \hline
        Model & Parameters (M) & Model Steps/sec (normalized) \\
        \hline
        Semla & 20 & 0.25 \\
        Megalodon Quick & 19 & 0.79 \\
        EQGAT-Diff & 12 & 1.00 \\
        Megalodon & 40 & 1.57 \\
        \hline
    \end{tabular}
    \label{tab:speed}
\end{table}

\begin{table}[!htbp]
    \centering
    
    \caption{Model Normalized Throughput (valid \& stable molecules/time)}
    \begin{tabular}{l|c|c|c|c}
        \hline
        Model & Parameters (M) & 30 Atoms & 100 Atoms & 120 Atoms \\
        \hline
        EQGAT-Diff & 12 & 1 & 1 & 1 \\
        Megalodon Quick & 19 & \textbf{1.21} & \textbf{3.9} & \textbf{44.0} \\
        Megalodon & 40 & 0.57 & 2.3 & 32.0 \\
        \hline
    \end{tabular}
    \label{tab:throughput}
\end{table}

In \tref{tab:speed}, we present the normalized inference speed of the models, measured in model steps per second relative to EQGAT-Diff. EQGAT-Diff serves as the baseline with a speed of 1.00. We present it this way as when using Flow Matching only 100 steps are needed no matter the architecture compared to diffusion's 500.

\tref{tab:throughput} shows the normalized throughput of the models, measured in valid and stable molecules per second for generating 100 molecules of different atom sizes. We exclude Semla from this table because it filters out all molecules larger than 80 atoms in their training data. Megalodon Quick significantly outperforms the other models, offering throughput that is up to 44 times faster than EQGAT-Diff (Table \ref{tab:throughput}). Megalodon also demonstrates throughput higher than that of EQGAT-Diff, particularly for molecules with 100 and 120 atoms.

\vspace{-1ex}
\subsection{Conditional Structure Generation}
\label{sec:rmsd}
Similar to the 3D molecule generation task, we use the GEOM-Drugs dataset to evaluate the conditional structure generation capabilities of our model. Given all unconditional 3DMG models are trained with independent noising of coordinates, atoms, and, in some cases, bonds, we want to evaluate how accurate the structural component is. We note this is something that is lacking from the existing prior benchmarks, as when generating novel de novo molecules, there is no ground truth structure to compare against. In the task of conditional structure generation, all models are given the molecule 2D graph (atom types, bonds) and asked to generate the 3D structure in which ground truth data exists.
Given \citet{MiDi} and \citet{torsional-diffusion} use different train/test splits, we evaluate all methods on the overlap of 200 held-out molecules, with all methods generating 43634 structures in total.
Due to the similarities with the baselines and its superior unconditional structure accuracy, we compare \model trained with diffusion against recent methods with public reproducible code.
\vspace{-1ex}
\paragraph{Problem setup.}
We report the average minimum RMSD (AMR) between ground truth and generated structures, and Coverage for Recall and Precision. Coverage is defined as the percentage of conformers with a minimum error under a specified AMR threshold. Recall matches each ground truth structure to its closest generated structure, and Precision measures the overall spatial accuracy of each generated structure.
Following \citet{torsional-diffusion}, we generate two times the number of ground truth structures for each molecule. More formally the precision metrics are defined, for $K=2L$, let $\{C^*_l\}_{l \in [1, L]}$ and $\{C_k\}_{k \in [1, K]}$ respectively be the sets of ground truth and generated structures:
\begin{equation}
\small
\begin{aligned}
&\text{COV-Prec.} := \frac{1}{K}\, \bigg\lvert\{ k \in [1..K]: \text{min}_{ l\in [1..L]} \rmsd(C_k, C^*_l) < \delta \} \bigg\rvert\\
&\text{AMR-Prec.} := \frac{1}{K} \sum_{k \in [1..K]}  \text{min}_{ l\in [1..L]} \rmsd(C_k, C^*_l)\\
\end{aligned}
\label{eq:amr}
\end{equation}
where $\delta$ is the coverage threshold. The recall metrics are obtained by swapping ground truth and generated conformers. 
\paragraph{Baselines}
We compare \model with EQGAT-Diff, GeoDiff~\citep{geodiff}, and TorsionalDiffusion~\citep{torsional-diffusion}.
For the unconditional 3DMG models, including \modelns, we prompt them with the ground truth atom types and bond types to guide the generation of the structure along the diffusion process. This is done by replacing the input and output with the fixed conditional data. We do this to assess what the model is actually learning across the multiple data domains. The central question being, is the model learning how to generate molecules over the spatial and discrete manifolds, or is it just learning how to copy snapshots of training-like data?

\paragraph{Analysis} We see in~\tref{tab:conf_gen} that EQGAT-diff is unable to generate any remotely valid structures. Even though all modalities are being denoised independently at different rates, the model cannot generate the structure given ground truth 2D molecule graphs. This is also seen during the sampling process, where diffusion models trained with similar denoising objectives as EQGAT-diff generate no bonds until the structure has seemingly converged. Therefore during most of the sampling process, the edge features which make up a large portion of the computational cost hold no value. 

In comparison, \model generates structures with competitive precision and recall by building a relationship between the discrete and continuous data directly in the training process described in~\sref{sec:methods}.
Half the time all data types are independently noised as normal with their respective time variables and schedules, the other half we only add noise to the structure. Therefore, our model learns to build a relationship between true 2D graphs and their 3D structure, as well as any interpolation between the three data tracks that are interpolated independently with different schedulers.

\begin{table}[H]
\centering
\caption{Quality of ML generated conformer ensembles for GEOM-DRUGS ($\delta=0.75$\AA) test set in terms of Coverage (\%) and Average RMSD (\AA). Bolded best, underlined second best.}
\label{tab:conf_gen}
\begin{tabular}{lc|cccc|cccc} \toprule
               & \multicolumn{1}{l|}{} & \multicolumn{4}{c|}{Recall} & \multicolumn{4}{c}{Precision}  \\
                 & \multicolumn{1}{l|}{} & \multicolumn{2}{c}{Coverage $\uparrow$} & \multicolumn{2}{c|}{AMR $\downarrow$} & \multicolumn{2}{c}{Coverage $\uparrow$} & \multicolumn{2}{c}{AMR $\downarrow$} \\
Method  &  & Mean & Med & Mean & Med & Mean & Med & Mean & Med \\ \midrule
\multicolumn{2}{l|}{GeoDiff} &  \textcolor{black}{42.1} & \textcolor{black}{37.8} & \textcolor{black}{0.835} & \textcolor{black}{0.809} & 24.9 & 14.5 & 1.136 & 1.090 \\
\multicolumn{2}{l|}{Tor. Diff.} & \textbf{75.3} & \textbf{82.3} &  \textbf{0.569}  &  \textbf{0.532} & \underline{56.5} & \underline{57.9} & \underline{0.778} & \underline{0.731} \\
\multicolumn{2}{l|}{EQGAT} &   0.8 & 0.0 & 2.790 & 2.847 & 0.1 & 0.0 & 3.754 & 3.771 \\
\multicolumn{2}{l|}{\modelns} & \underline{71.4} & \underline{75.0} & \underline{0.573} & \underline{0.557} & \textbf{61.2} & \textbf{63.1} & \textbf{0.719} & \textbf{0.696} \\
\bottomrule
\end{tabular}
\vspace{-3ex}
\end{table}

\model demonstrates that its unconditional discrete diffusion objective is crucial for its conditional performance.
In other words, the discrete diffusion training objective improves the conditional continuous generative performance. This is evident in the comparison between GeoDiff and \modelns. GeoDiff is trained on the same conditional Euclidean structure objective as \model(with similar EGNN-based architecture) with 10x more diffusion steps, with both models taking in identical inputs. We see that since \model is able to generate molecules from pure noise, it better learns structure and as a result can be prompted to generate accurate structures.

Interestingly, compared to Torsional Diffusion, which initializes the 3D structure with an expensive RDKit approximation to establish all bond lengths and angles and then only modifies the dihedral angles, we see quite competitive performance. 
Before, it was understood that by restricting the degrees of freedom with good RDKit structures, the performance jump from GeoDiff to Torsional Diffusion was observed. Now we see that with the same euclidean diffusion process, similar accuracy improvements can be gained by learning how to generate accurate discrete molecule topology via discrete diffusion.
We want to note that there have been recent advances on top of Torsional Diffusion~\citep{particleguidance} and other conformer-focused models that are not public~\citep{swallow}. We use this benchmark more to analyze the underlying multi-modal diffusion objective and focus on the underlying model comparisons. 
\model is not a conformer generation model but a molecule generation model capable of de novo and conditional design.
Overall, \model shows how independent time interpolation and discrete diffusion create the ability for the model to be prompted or guided with a desired 2D topology to generate accurate 3D structures.

\vspace{-1ex}
\subsection{Unconditional Structure-based Energy Benchmarks}
\label{sec:xtb}

\begin{table*}[h]
\caption{xTB Relaxation Error: Length \AA, angles degrees, energy kcal/mol. These metrics are taken over the valid molecules from~\tref{tab:3dmg}. Methods are grouped by model type: diffusion (500 steps) and flow matching (100 steps)}
\label{tab:xtb}
\centering
\begin{tabular}{l c | c c c | c c c}
\toprule
Model &  & Bond Length & Bond Angles & Dihedral & Median $\Delta E_{\text{relax}}$ & Mean $\Delta E_{\text{relax}}$ \\ \midrule
\multicolumn{2}{l|}{GEOM-Drugs} & 0.0000 & 0.00 & 7.2e-3 & 0.00 & 1.0e-3 \\
\midrule
\multicolumn{2}{l|}{EQGAT-diff} & 0.0076 & 0.95 & 7.98 & 6.36 & 11.06 \\

\multicolumn{2}{l|}{\modelns-quick} & 0.0085 & 0.88 & 7.28 & 5.78 & 9.74  \\
\multicolumn{2}{l|}{\modelns} & \textbf{0.0061} & \textbf{0.66} & \textbf{5.42} & \textbf{3.17} & \textbf{5.71}   \\
\midrule
\multicolumn{2}{l|}{SemlaFlow} & 0.0309 & 2.03 & 6.01 & 32.96 & 93.13  \\
\multicolumn{2}{l|}{\modelns-flow} & \textbf{0.0225} & \textbf{1.59} & \textbf{5.49} & \textbf{20.86} & \textbf{46.86} \\
\bottomrule
\end{tabular}
\vspace{-2ex}
\end{table*}
\paragraph{Problem setup} Each ground truth structure in GEOM dataset represents a low-energy conformer within its ensemble, highlighting two key aspects. First, these molecules are local minima on the GFN2-xTB potential energy surface. Second, their energies are lower compared to other conformations sampled in the ensemble. Previously, these quantities have not been thoroughly evaluated for generated molecules. To address this gap, we directly measure how closely a generated molecule approximates its nearest local minimum (i.e., its relaxed structure). We measure the energy difference between the initial generated structure and its relaxed counterpart, as well as structural changes in bond lengths, bond angles, and dihedral (torsion) angles. This approach allows us to evaluate the ability of generative models to produce molecules that are true local minima, facilitating faster ranking of generated structures without additional minimization steps.

\vspace{-1ex}
\paragraph{Analysis}
In~\tref{tab:xtb} we see that both diffusion and flow matching, \model is better than its prior counterparts. Overall, \model trained with diffusion performs best with roughly 2-10x lower median energy when compared to prior generative models. Notably, our model's median relaxation energy difference $\Delta E_{\text{relax}}$ is around $3$ $\text{kcal/mol}$, which approaches the thermally relevant interval of $2.5$ $\text{kcal/mol}$~\citep{geom}. \model is the first method to achieve such proximity to this thermodynamic threshold, marking a significant milestone in 3D molecular generation. For more details, please see Appendix~\sref{sec:energy-app}.

We note that while the loss function between FM and diffusion is identical in this instance, we see both flow models have an order of magnitude larger bond angle error, which translates to a similar energy performance gap. The xTB energy function is highly sensitive to bond lengths; small deviations in bond lengths can lead to significant increases in energy due to the steepness of the potential energy surface in these dimensions. A precise representation of bond lengths is crucial because inaccuracies directly impact the calculated energy, making bond length errors a primary contributor to higher relaxation energies in flow models.
We hypothesize that since the Flow models scale the input structures to have a variance of 1 to make matching the Gaussian prior easier, we lower the local spatial precision necessary for bond length and angle generation~\citep{semla}. 

\vspace{-1ex}
\section{Conclusions}
\vspace{-1ex}
\model enables the accurate generation of de novo 3D molecules with both diffusion and flow matching.
We show with a scalable augmented transformer architecture that significant improvements are gained, especially when generating outside the region of support for the training distribution as it pertains to molecule sizes.
\model demonstrates the ability to achieve great accuracy in conditional structure generation due to being trained to generate complete molecules from scratch.
We also introduce more interpretable quantum mechanical energy benchmarks that are grounded in the original creation of the GEOM Drugs dataset.
Overall, we explore the similarities and differences between flow matching and diffusion while improving 3D molecule design.

\section{Acknowledgement}
O.I. acknowledges support by the NSF grant CHE-2154447. This work used Expanse at SDSC and Delta at NCSA through allocation CHE200122 from the Advanced Cyberinfrastructure Coordination Ecosystem: Services \& Support (ACCESS) program, which is supported by NSF grants \#2138259, \#2138286, \#2138307, \#2137603, and \#2138296.

\newpage
\newpage
\bibliography{paper}

\begin{thebibliography}{47}
\providecommand{\natexlab}[1]{#1}
\providecommand{\url}[1]{\texttt{#1}}
\expandafter\ifx\csname urlstyle\endcsname\relax
  \providecommand{\doi}[1]{doi: #1}\else
  \providecommand{\doi}{doi: \begingroup \urlstyle{rm}\Url}\fi

\bibitem[Abramson et~al.(2024)Abramson, Adler, Dunger, Evans, Green, Pritzel, Ronneberger, Willmore, Ballard, Bambrick, Bodenstein, Evans, Hung, O'Neill, Reiman, Tunyasuvunakool, Wu, {\v{Z}}emgulyt{\.{e}}, Arvaniti, Beattie, Bertolli, Bridgland, Cherepanov, Congreve, Cowen-Rivers, Cowie, Figurnov, Fuchs, Gladman, Jain, Khan, Low, Perlin, Potapenko, Savy, Singh, Stecula, Thillaisundaram, Tong, Yakneen, Zhong, Zielinski, {\v{Z}}{\'i}dek, Bapst, Kohli, Jaderberg, Hassabis, and Jumper]{af3}
Josh Abramson, Jonas Adler, Jack Dunger, Richard Evans, Tim Green, Alexander Pritzel, Olaf Ronneberger, Lindsay Willmore, Andrew~J. Ballard, Joshua Bambrick, Sebastian~W. Bodenstein, David~A. Evans, Chia-Chun Hung, Michael O'Neill, David Reiman, Kathryn Tunyasuvunakool, Zachary Wu, Akvil{\.{e}} {\v{Z}}emgulyt{\.{e}}, Eirini Arvaniti, Charles Beattie, Ottavia Bertolli, Alex Bridgland, Alexey Cherepanov, Miles Congreve, Alexander~I. Cowen-Rivers, Andrew Cowie, Michael Figurnov, Fabian~B. Fuchs, Hannah Gladman, Rishub Jain, Yousuf~A. Khan, Caroline M.~R. Low, Kuba Perlin, Anna Potapenko, Pascal Savy, Sukhdeep Singh, Adrian Stecula, Ashok Thillaisundaram, Catherine Tong, Sergei Yakneen, Ellen~D. Zhong, Michal Zielinski, Augustin {\v{Z}}{\'i}dek, Victor Bapst, Pushmeet Kohli, Max Jaderberg, Demis Hassabis, and John~M. Jumper.
\newblock Accurate structure prediction of biomolecular interactions with alphafold 3.
\newblock \emph{Nature}, 630\penalty0 (8016):\penalty0 493--500, Jun 2024.
\newblock ISSN 1476-4687.
\newblock \doi{10.1038/s41586-024-07487-w}.
\newblock URL \url{https://doi.org/10.1038/s41586-024-07487-w}.

\bibitem[Albergo et~al.(2023)Albergo, Boffi, and Vanden-Eijnden]{albergo2023stochastic}
Michael~S. Albergo, Nicholas~M. Boffi, and Eric Vanden-Eijnden.
\newblock Stochastic interpolants: A unifying framework for flows and diffusions, 2023.
\newblock URL \url{https://arxiv.org/abs/2303.08797}.

\bibitem[Austin et~al.(2021)Austin, Johnson, Ho, Tarlow, and Van Den~Berg]{d3pm}
Jacob Austin, Daniel~D Johnson, Jonathan Ho, Daniel Tarlow, and Rianne Van Den~Berg.
\newblock Structured denoising diffusion models in discrete state-spaces.
\newblock \emph{Advances in Neural Information Processing Systems}, 34:\penalty0 17981--17993, 2021.

\bibitem[Axelrod \& G{\'o}mez-Bombarelli(2022)Axelrod and G{\'o}mez-Bombarelli]{geom}
Simon Axelrod and Rafael G{\'o}mez-Bombarelli.
\newblock Geom, energy-annotated molecular conformations for property prediction and molecular generation.
\newblock \emph{Scientific Data}, 9\penalty0 (1):\penalty0 185, 2022.
\newblock \doi{10.1038/s41597-022-01288-4}.
\newblock URL \url{https://doi.org/10.1038/s41597-022-01288-4}.

\bibitem[Bannwarth et~al.(2019)Bannwarth, Ehlert, and Grimme]{xtb}
Christoph Bannwarth, Sebastian Ehlert, and Stefan Grimme.
\newblock Gfn2-xtb---an accurate and broadly parametrized self-consistent tight-binding quantum chemical method with multipole electrostatics and density-dependent dispersion contributions.
\newblock \emph{Journal of Chemical Theory and Computation}, 15\penalty0 (3):\penalty0 1652--1671, Mar 2019.

\bibitem[B{\'e}k{\'e}s et~al.(2022)B{\'e}k{\'e}s, Langley, and Crews]{protac}
Mikl{\'o}s B{\'e}k{\'e}s, David~R. Langley, and Craig~M. Crews.
\newblock Protac targeted protein degraders: the past is prologue.
\newblock \emph{Nature Reviews Drug Discovery}, 21\penalty0 (3):\penalty0 181--200, 2022.
\newblock \doi{10.1038/s41573-021-00371-6}.
\newblock URL \url{https://doi.org/10.1038/s41573-021-00371-6}.

\bibitem[Brown et~al.(2020)Brown, Mann, Ryder, Subbiah, Kaplan, Dhariwal, Neelakantan, Shyam, Sastry, Askell, Agarwal, Herbert-Voss, Krueger, Henighan, Child, Ramesh, Ziegler, Wu, Winter, Hesse, Chen, Sigler, Litwin, Gray, Chess, Clark, Berner, McCandlish, Radford, Sutskever, and Amodei]{gpt3}
Tom~B. Brown, Benjamin Mann, Nick Ryder, Melanie Subbiah, Jared Kaplan, Prafulla Dhariwal, Arvind Neelakantan, Pranav Shyam, Girish Sastry, Amanda Askell, Sandhini Agarwal, Ariel Herbert-Voss, Gretchen Krueger, Tom Henighan, Rewon Child, Aditya Ramesh, Daniel~M. Ziegler, Jeffrey Wu, Clemens Winter, Christopher Hesse, Mark Chen, Eric Sigler, Mateusz Litwin, Scott Gray, Benjamin Chess, Jack Clark, Christopher Berner, Sam McCandlish, Alec Radford, Ilya Sutskever, and Dario Amodei.
\newblock Language models are few-shot learners, 2020.

\bibitem[Campbell et~al.(2024)Campbell, Yim, Barzilay, Rainforth, and Jaakkola]{multiflow}
Andrew Campbell, Jason Yim, Regina Barzilay, Tom Rainforth, and Tommi Jaakkola.
\newblock Generative flows on discrete state-spaces: Enabling multimodal flows with applications to protein co-design.
\newblock \emph{arXiv preprint arXiv:2402.04997}, 2024.

\bibitem[Chen et~al.(2022)Chen, Zhang, and Hinton]{selfconditioning}
Ting Chen, Ruixiang Zhang, and Geoffrey Hinton.
\newblock Analog bits: Generating discrete data using diffusion models with self-conditioning.
\newblock \emph{arXiv preprint arXiv:2208.04202}, 2022.

\bibitem[Corso et~al.(2023{\natexlab{a}})Corso, Stärk, Jing, Barzilay, and Jaakkola]{diffdock}
Gabriele Corso, Hannes Stärk, Bowen Jing, Regina Barzilay, and Tommi Jaakkola.
\newblock Diffdock: Diffusion steps, twists, and turns for molecular docking.
\newblock In \emph{International Conference on Learning Representations (ICLR)}, 2023{\natexlab{a}}.

\bibitem[Corso et~al.(2023{\natexlab{b}})Corso, Xu, De~Bortoli, Barzilay, and Jaakkola]{particleguidance}
Gabriele Corso, Yilun Xu, Valentin De~Bortoli, Regina Barzilay, and Tommi Jaakkola.
\newblock Particle guidance: non-iid diverse sampling with diffusion models.
\newblock \emph{arXiv preprint arXiv:2310.13102}, 2023{\natexlab{b}}.

\bibitem[Durairaj et~al.(2024)Durairaj, Adeshina, Cao, Zhang, Oleinikovas, Duignan, McClure, Robin, Studer, Kovtun, Rossi, Zhou, Veccham, Isert, Peng, Sundareson, Akdel, Corso, St{\"a}rk, Tauriello, Carpenter, Bronstein, Kucukbenli, Schwede, and Naef]{plinder}
Janani Durairaj, Yusuf Adeshina, Zhonglin Cao, Xuejin Zhang, Vladas Oleinikovas, Thomas Duignan, Zachary McClure, Xavier Robin, Gabriel Studer, Daniel Kovtun, Emanuele Rossi, Guoqing Zhou, Srimukh Veccham, Clemens Isert, Yuxing Peng, Prabindh Sundareson, Mehmet Akdel, Gabriele Corso, Hannes St{\"a}rk, Gerardo Tauriello, Zachary Carpenter, Michael Bronstein, Emine Kucukbenli, Torsten Schwede, and Luca Naef.
\newblock Plinder: The protein-ligand interactions dataset and evaluation resource.
\newblock \emph{bioRxiv}, 2024.
\newblock \doi{10.1101/2024.07.17.603955}.
\newblock URL \url{https://www.biorxiv.org/content/early/2024/07/19/2024.07.17.603955.1}.

\bibitem[Esser et~al.(2024)Esser, Kulal, Blattmann, Entezari, M{\"u}ller, Saini, Levi, Lorenz, Sauer, Boesel, Podell, Dockhorn, English, and Rombach]{esser2024scaling}
Patrick Esser, Sumith Kulal, Andreas Blattmann, Rahim Entezari, Jonas M{\"u}ller, Harry Saini, Yam Levi, Dominik Lorenz, Axel Sauer, Frederic Boesel, Dustin Podell, Tim Dockhorn, Zion English, and Robin Rombach.
\newblock Scaling rectified flow transformers for high-resolution image synthesis.
\newblock In \emph{Forty-first International Conference on Machine Learning}, 2024.
\newblock URL \url{https://openreview.net/forum?id=FPnUhsQJ5B}.

\bibitem[Foloppe \& Chen(2019)Foloppe and Chen]{foloppe2019energy}
Nicolas Foloppe and I-Jen Chen.
\newblock Energy windows for computed compound conformers: covering artefacts or truly large reorganization energies?
\newblock \emph{Future Medicinal Chemistry}, 11\penalty0 (2):\penalty0 97--118, 2019.

\bibitem[Gómez-Bombarelli et~al.(2018)Gómez-Bombarelli, Wei, Duvenaud, Hernández-Lobato, Sánchez-Lengeling, Sheberla, Aguilera-Iparraguirre, Hirzel, Adams, and Aspuru-Guzik]{gomez-vae}
Rafael Gómez-Bombarelli, Jennifer~N. Wei, David Duvenaud, José~Miguel Hernández-Lobato, Benjamín Sánchez-Lengeling, Dennis Sheberla, Jorge Aguilera-Iparraguirre, Timothy~D. Hirzel, Ryan~P. Adams, and Alán Aspuru-Guzik.
\newblock Automatic chemical design using a data-driven continuous representation of molecules.
\newblock \emph{ACS Central Science}, 4\penalty0 (2):\penalty0 268--276, 2018.
\newblock \doi{10.1021/acscentsci.7b00572}.
\newblock URL \url{https://doi.org/10.1021/acscentsci.7b00572}.
\newblock PMID: 29532027.

\bibitem[Halgren(1996)]{mmff}
Thomas~A. Halgren.
\newblock Merck molecular force field. i. basis, form, scope, parameterization, and performance of mmff94.
\newblock \emph{Journal of Computational Chemistry}, 17\penalty0 (5-6):\penalty0 490--519, 1996.
\newblock \doi{https://doi.org/10.1002/(SICI)1096-987X(199604)17:5/6<490::AID-JCC1>3.0.CO;2-P}.

\bibitem[Hayes et~al.(2024)Hayes, Rao, Akin, Sofroniew, Oktay, Lin, Verkuil, Tran, Deaton, Wiggert, Badkundri, Shafkat, Gong, Derry, Molina, Thomas, Khan, Mishra, Kim, Bartie, Nemeth, Hsu, Sercu, Candido, and Rives]{esm3}
Thomas Hayes, Roshan Rao, Halil Akin, Nicholas~J. Sofroniew, Deniz Oktay, Zeming Lin, Robert Verkuil, Vincent~Q. Tran, Jonathan Deaton, Marius Wiggert, Rohil Badkundri, Irhum Shafkat, Jun Gong, Alexander Derry, Raul~S. Molina, Neil Thomas, Yousuf Khan, Chetan Mishra, Carolyn Kim, Liam~J. Bartie, Matthew Nemeth, Patrick~D. Hsu, Tom Sercu, Salvatore Candido, and Alexander Rives.
\newblock Simulating 500 million years of evolution with a language model.
\newblock \emph{bioRxiv}, 2024.
\newblock \doi{10.1101/2024.07.01.600583}.
\newblock URL \url{https://www.biorxiv.org/content/early/2024/07/02/2024.07.01.600583}.

\bibitem[Henry et~al.(2020)Henry, Dachapally, Pawar, and Chen]{qk_norm}
Alex Henry, Prudhvi~Raj Dachapally, Shubham~Shantaram Pawar, and Yuxuan Chen.
\newblock Query-key normalization for transformers.
\newblock In \emph{Findings of the Association for Computational Linguistics: EMNLP 2020}, pp.\  4246–4253. Association for Computational Linguistics, 2020.
\newblock \doi{10.18653/v1/2020.findings-emnlp.379}.
\newblock URL \url{http://dx.doi.org/10.18653/v1/2020.findings-emnlp.379}.

\bibitem[Ho et~al.(2020)Ho, Jain, and Abbeel]{DDPM}
Jonathan Ho, Ajay Jain, and Pieter Abbeel.
\newblock Denoising diffusion probabilistic models.
\newblock \emph{arXiv preprint arxiv:2006.11239}, 2020.

\bibitem[Hoogeboom et~al.(2022)Hoogeboom, Satorras, Vignac, and Welling]{EDM}
Emiel Hoogeboom, V{\i}ctor~Garcia Satorras, Cl{\'e}ment Vignac, and Max Welling.
\newblock Equivariant diffusion for molecule generation in 3d.
\newblock In \emph{International conference on machine learning}, pp.\  8867--8887. PMLR, 2022.

\bibitem[Irwin et~al.(2024)Irwin, Tibo, Janet, and Olsson]{semla}
Ross Irwin, Alessandro Tibo, Jon~Paul Janet, and Simon Olsson.
\newblock Efficient 3d molecular generation with flow matching and scale optimal transport, 2024.

\bibitem[Jing et~al.(2022)Jing, Corso, Chang, Barzilay, and Jaakkola]{torsional-diffusion}
Bowen Jing, Gabriele Corso, Jeffrey Chang, Regina Barzilay, and Tommi Jaakkola.
\newblock Torsional diffusion for molecular conformer generation.
\newblock \emph{arXiv preprint arXiv:2206.01729}, 2022.

\bibitem[Le et~al.(2024)Le, Cremer, Noe, Clevert, and Sch{\"u}tt]{eqgat}
Tuan Le, Julian Cremer, Frank Noe, Djork-Arn{\'e} Clevert, and Kristof~T Sch{\"u}tt.
\newblock Navigating the design space of equivariant diffusion-based generative models for de novo 3d molecule generation.
\newblock In \emph{The Twelfth International Conference on Learning Representations}, 2024.
\newblock URL \url{https://openreview.net/forum?id=kzGuiRXZrQ}.

\bibitem[Lin et~al.(2023)Lin, Akin, Rao, Hie, Zhu, Lu, Smetanin, Verkuil, Kabeli, Shmueli, et~al.]{esm}
Zeming Lin, Halil Akin, Roshan Rao, Brian Hie, Zhongkai Zhu, Wenting Lu, Nikita Smetanin, Robert Verkuil, Ori Kabeli, Yaniv Shmueli, et~al.
\newblock Evolutionary-scale prediction of atomic-level protein structure with a language model.
\newblock \emph{Science}, 379\penalty0 (6637):\penalty0 1123--1130, 2023.

\bibitem[Lipman et~al.(2023)Lipman, Chen, Ben-Hamu, Nickel, and Le]{lipman2023flow}
Yaron Lipman, Ricky T.~Q. Chen, Heli Ben-Hamu, Maximilian Nickel, and Matthew Le.
\newblock Flow matching for generative modeling.
\newblock In \emph{The Eleventh International Conference on Learning Representations}, 2023.
\newblock URL \url{https://openreview.net/forum?id=PqvMRDCJT9t}.

\bibitem[Ma et~al.(2024)Ma, Goldstein, Albergo, Boffi, Vanden-Eijnden, and Xie]{sit}
Nanye Ma, Mark Goldstein, Michael~S. Albergo, Nicholas~M. Boffi, Eric Vanden-Eijnden, and Saining Xie.
\newblock Sit: Exploring flow and diffusion-based generative models with scalable interpolant transformers, 2024.
\newblock URL \url{https://arxiv.org/abs/2401.08740}.

\bibitem[O'Boyle et~al.(2011)O'Boyle, Banck, James, Morley, Vandermeersch, and Hutchison]{openbable}
Noel~M. O'Boyle, Michael Banck, Craig~A. James, Chris Morley, Tim Vandermeersch, and Geoffrey~R. Hutchison.
\newblock Open babel: An open chemical toolbox.
\newblock \emph{Journal of Cheminformatics}, 3\penalty0 (1):\penalty0 33, Oct 2011.
\newblock ISSN 1758-2946.
\newblock \doi{10.1186/1758-2946-3-33}.
\newblock URL \url{https://doi.org/10.1186/1758-2946-3-33}.

\bibitem[Peebles \& Xie(2022)Peebles and Xie]{DiT}
William Peebles and Saining Xie.
\newblock Scalable diffusion models with transformers.
\newblock \emph{arXiv preprint arXiv:2212.09748}, 2022.

\bibitem[Peng et~al.(2023)Peng, Guan, Liu, and Ma]{MolDiff}
Xingang Peng, Jiaqi Guan, Qiang Liu, and Jianzhu Ma.
\newblock {M}ol{D}iff: Addressing the atom-bond inconsistency problem in 3{D} molecule diffusion generation.
\newblock In Andreas Krause, Emma Brunskill, Kyunghyun Cho, Barbara Engelhardt, Sivan Sabato, and Jonathan Scarlett (eds.), \emph{Proceedings of the 40th International Conference on Machine Learning}, volume 202 of \emph{Proceedings of Machine Learning Research}, pp.\  27611--27629. PMLR, 23--29 Jul 2023.
\newblock URL \url{https://proceedings.mlr.press/v202/peng23b.html}.

\bibitem[Pinheiro et~al.(2024)Pinheiro, Rackers, Kleinhenz, Maser, Mahmood, Watkins, Ra, Sresht, and Saremi]{voxmol}
Pedro~O. Pinheiro, Joshua Rackers, Joseph Kleinhenz, Michael Maser, Omar Mahmood, Andrew~Martin Watkins, Stephen Ra, Vishnu Sresht, and Saeed Saremi.
\newblock 3d molecule generation by denoising voxel grids, 2024.
\newblock URL \url{https://arxiv.org/abs/2306.07473}.

\bibitem[Pracht et~al.(2024)Pracht, Grimme, Bannwarth, Bohle, Ehlert, Feldmann, Gorges, M{\"u}ller, Neudecker, Plett, et~al.]{crest}
Philipp Pracht, Stefan Grimme, Christoph Bannwarth, Fabian Bohle, Sebastian Ehlert, Gereon Feldmann, Johannes Gorges, Marcel M{\"u}ller, Tim Neudecker, Christoph Plett, et~al.
\newblock Crest—a program for the exploration of low-energy molecular chemical space.
\newblock \emph{The Journal of Chemical Physics}, 160\penalty0 (11), 2024.

\bibitem[Reidenbach(2024)]{evosbdd}
Danny Reidenbach.
\newblock Evo{SBDD}: Latent evolution for accurate and efficient structure-based drug design.
\newblock In \emph{ICLR 2024 Workshop on Machine Learning for Genomics Explorations}, 2024.
\newblock URL \url{https://openreview.net/forum?id=sLhUNz0uTz}.

\bibitem[Satorras et~al.(2021)Satorras, Hoogeboom, and Welling]{egnn}
V{\i}ctor~Garcia Satorras, Emiel Hoogeboom, and Max Welling.
\newblock E (n) equivariant graph neural networks.
\newblock In \emph{International conference on machine learning}, pp.\  9323--9332. PMLR, 2021.

\bibitem[Schneuing et~al.(2022)Schneuing, Du, Harris, Jamasb, Igashov, Du, Blundell, Li{\'o}, Gomes, Welling, Bronstein, and Correia]{diffsbdd}
Arne Schneuing, Yuanqi Du, Charles Harris, Arian Jamasb, Ilia Igashov, Weitao Du, Tom Blundell, Pietro Li{\'o}, Carla Gomes, Max Welling, Michael Bronstein, and Bruno Correia.
\newblock Structure-based drug design with equivariant diffusion models.
\newblock \emph{arXiv preprint arXiv:2210.13695}, 2022.

\bibitem[Song et~al.(2021)Song, Sohl-Dickstein, Kingma, Kumar, Ermon, and Poole]{song2020score}
Yang Song, Jascha Sohl-Dickstein, Diederik~P Kingma, Abhishek Kumar, Stefano Ermon, and Ben Poole.
\newblock {Score-Based Generative Modeling through Stochastic Differential Equations}.
\newblock In \emph{International Conference on Learning Representations (ICLR)}, 2021.

\bibitem[Song et~al.(2023)Song, Gong, Xu, Cao, Lan, Ermon, Zhou, and Ma]{EquiFM}
Yuxuan Song, Jingjing Gong, Minkai Xu, Ziyao Cao, Yanyan Lan, Stefano Ermon, Hao Zhou, and Wei-Ying Ma.
\newblock Equivariant flow matching with hybrid probability transport for 3d molecule generation.
\newblock In \emph{Thirty-seventh Conference on Neural Information Processing Systems}, 2023.
\newblock URL \url{https://openreview.net/forum?id=hHUZ5V9XFu}.

\bibitem[Song et~al.(2024)Song, Gong, Qu, Zhou, Zheng, Liu, and Ma]{GeoBFN}
Yuxuan Song, Jingjing Gong, Yanru Qu, Hao Zhou, Mingyue Zheng, Jingjing Liu, and Wei-Ying Ma.
\newblock Unified generative modeling of 3d molecules via bayesian flow networks, 2024.
\newblock URL \url{https://arxiv.org/abs/2403.15441}.

\bibitem[Stärk et~al.(2024)Stärk, Jing, Barzilay, and Jaakkola]{harmonicflow}
Hannes Stärk, Bowen Jing, Regina Barzilay, and Tommi Jaakkola.
\newblock Harmonic self-conditioned flow matching for multi-ligand docking and binding site design, 2024.
\newblock URL \url{https://arxiv.org/abs/2310.05764}.

\bibitem[Tong et~al.(2023)Tong, Malkin, Huguet, Zhang, Rector-Brooks, Fatras, Wolf, and Bengio]{tong2023improving}
Alexander Tong, Nikolay Malkin, Guillaume Huguet, Yanlei Zhang, Jarrid Rector-Brooks, Kilian Fatras, Guy Wolf, and Yoshua Bengio.
\newblock Improving and generalizing flow-based generative models with minibatch optimal transport.
\newblock \emph{arXiv preprint arXiv:2302.00482}, 2023.

\bibitem[Vaswani(2017)]{attention}
A~Vaswani.
\newblock Attention is all you need.
\newblock \emph{Advances in Neural Information Processing Systems}, 2017.

\bibitem[Vignac et~al.(2023)Vignac, Osman, Toni, and Frossard]{MiDi}
Clement Vignac, Nagham Osman, Laura Toni, and Pascal Frossard.
\newblock Midi: Mixed graph and 3d denoising diffusion for molecule generation.
\newblock \emph{arXiv preprint arXiv:2302.09048}, 2023.

\bibitem[Vincent(2011)]{vincent2011}
Pascal Vincent.
\newblock A connection between score matching and denoising autoencoders.
\newblock \emph{Neural Computation}, 23\penalty0 (7):\penalty0 1661--1674, 2011.

\bibitem[Walters(2024)]{pat_walters_blog}
Pat Walters.
\newblock Generative molecular design isn't as easy as people make it look, May 2024.
\newblock URL \url{https://practicalcheminformatics.blogspot.com/2024/05/generative-molecular-design-isnt-as.html}.

\bibitem[Wang et~al.(2023)Wang, Elhag, Jaitly, Susskind, and Bautista]{swallow}
Yuyang Wang, Ahmed~A. Elhag, Navdeep Jaitly, Joshua~M. Susskind, and Miguel~Angel Bautista.
\newblock Swallowing the bitter pill: Simplified scalable conformer generation, 2023.

\bibitem[Xu et~al.(2022)Xu, Yu, Song, Shi, Ermon, and Tang]{geodiff}
Minkai Xu, Lantao Yu, Yang Song, Chence Shi, Stefano Ermon, and Jian Tang.
\newblock Geodiff: A geometric diffusion model for molecular conformation generation.
\newblock In \emph{International Conference on Learning Representations}, 2022.
\newblock URL \url{https://openreview.net/forum?id=PzcvxEMzvQC}.

\bibitem[Xu et~al.(2023)Xu, Powers, Dror, Ermon, and Leskovec]{geoldm}
Minkai Xu, Alexander Powers, Ron Dror, Stefano Ermon, and Jure Leskovec.
\newblock Geometric latent diffusion models for 3d molecule generation, 2023.
\newblock URL \url{https://arxiv.org/abs/2305.01140}.

\bibitem[Yim et~al.(2023)Yim, Trippe, De~Bortoli, Mathieu, Doucet, Barzilay, and Jaakkola]{FrameDiff}
Jason Yim, Brian~L Trippe, Valentin De~Bortoli, Emile Mathieu, Arnaud Doucet, Regina Barzilay, and Tommi Jaakkola.
\newblock Se (3) diffusion model with application to protein backbone generation.
\newblock \emph{arXiv preprint arXiv:2302.02277}, 2023.

\end{thebibliography}
\bibliographystyle{iclr2025_conference}

\newpage
\appendix
\addcontentsline{toc}{section}{Appendix}
\part{Appendix} 
\parttoc 
\begin{figure}[ht]
    \centering
    \includegraphics[width=0.7\textwidth]{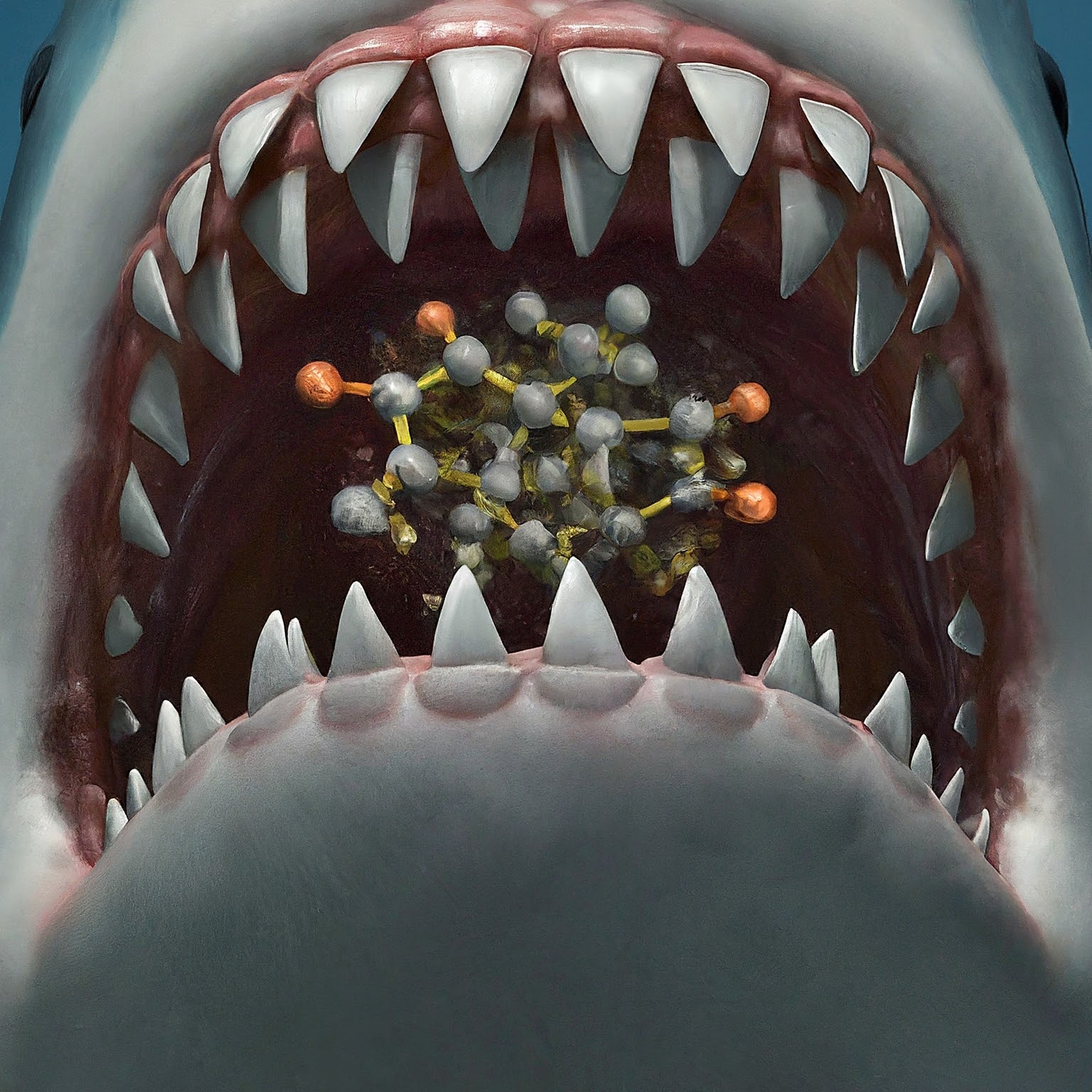}  
    \caption{\model molecule generation dynamics generated with Imagen 2}
    \label{fig:logo}
\end{figure}

\section{Equating Continuous Gaussian Diffusion and Flow Matching}
\label{sec:app-diff}

A part of our work was to explore when to use diffusion versus flow matching and what the empirical differences are. We show below that from a training perspective in the continuous domain, they can be made equivalent.

It can be shown that this objective under the Gaussian setting is a time-dependent scalar multiple of the standard denoising objective explored in~\citet{DDPM}.
Let's insert \eqref{eq:interpolant_b} into the flow matching objective
\begin{equation}
\begin{split}
    \mathcal{L}_\textrm{CFM}(\theta)&=\mathbb{E}_{t,\boldsymbol{\epsilon}\sim\mathcal{N}(\boldsymbol{\epsilon};\mathbf{0},\mI),\rvx_1\sim p_\textrm{data}(\rvx_1)}||\rvv_\theta(t,\rvx_t)-\dot{\alpha}(t)\boldsymbol{\epsilon} - \frac{\dot{\beta}(t)}{\beta(t)}(\rvx_t-\alpha(t)\boldsymbol{\epsilon})||^2.
\end{split}
\end{equation}
where the dot notation denotes the partial time derivative.

Now we see that we can construct an objective that is similar to the ``noise prediction'' objective that is used in diffusion models:
\begin{equation} \label{eq:cfm_reparam}
\begin{split}
    \mathcal{L}_\textrm{CFM}(\theta)&=\mathbb{E}_{t,\boldsymbol{\epsilon}\sim\mathcal{N}(\boldsymbol{\epsilon};\mathbf{0},\mI),\rvx_1\sim p_\textrm{data}(\rvx_1)}||\rvv_\theta(t,\rvx_t)-\dot{\alpha}(t)\boldsymbol{\epsilon} - \frac{\dot{\beta}(t)}{\beta(t)}(\rvx_t-\alpha(t)\boldsymbol{\epsilon})||^2 \\
    &=\mathbb{E}_{t,\boldsymbol{\epsilon}\sim\mathcal{N}(\boldsymbol{\epsilon};\mathbf{0},\mI),\rvx_1\sim p_\textrm{data}(\rvx_1)}||\rvv_\theta(t,\rvx_t)- \frac{\dot{\beta}(t)}{\beta(t)}\rvx_t  -\underbrace{(\dot{\alpha}(t) - \frac{\dot{\beta}(t)}{\beta(t)}\alpha(t))}_{=:s(t)}\boldsymbol{\epsilon}||^2 \\
    &=\mathbb{E}_{t,\boldsymbol{\epsilon}\sim\mathcal{N}(\boldsymbol{\epsilon};\mathbf{0},\mI),\rvx_1\sim p_\textrm{data}(\rvx_1)}s^2(t)||\underbrace{\frac{1}{s(t)}\left(\rvv_\theta(t,\rvx_t)- \frac{\dot{\beta}(t)}{\beta(t)}\rvx_t\right)}_{=:\boldsymbol{\epsilon}_\theta(t,\rvx_t)}  -\boldsymbol{\epsilon}||^2 \\
    &=\mathbb{E}_{t,\boldsymbol{\epsilon}\sim\mathcal{N}(\boldsymbol{\epsilon};\mathbf{0},\mI),\rvx_1\sim p_\textrm{data}(\rvx_1)}s^2(t)||\boldsymbol{\epsilon}_\theta(t,\rvx_t)  -\boldsymbol{\epsilon}||^2.
\end{split}
\end{equation}
We see that the resulting mean squared error of noise prediction is the original core loss derived in~\citet{DDPM}. This allows us to choose time-dependent scalars via the time distribution itself or the noise or variance schedule to equate the CFM and Diffusion objectives.

In the generative modeling case, we interpolate between a data distribution and a Gaussian density, meaning all data-conditional paths are Gaussian. In that special case, we can, in fact, easily extract the score function from the regular flow matching objective, and we get stochastic sampling for free. We know that $\rvx_t\sim p(\rvx_t|\rvx_1)$ follows Gaussian probability paths. Based on \eqref{eq:interpolant}, we know that
\begin{equation}
    \rvx_t \sim p(\rvx_t|\rvx_1) = \mathcal{N}(\rvx_t;\beta(t)\rvx_1, \alpha^2(t)\mI).
\end{equation}
Let's calculate the score:
\begin{equation}
\begin{split}
    \nabla_{\rvx_t}\log p(\rvx_t|\rvx_1)&=-\nabla_{\rvx_t}\frac{(\rvx_t-\beta(t)\rvx_1)^2}{2\alpha^2(t)} \\
    & = - \frac{\rvx_t-\beta(t)\rvx_1}{\alpha^2(t)} \\
    & = -\frac{\boldsymbol{\epsilon}}{\alpha(t)},
\end{split}
\end{equation}
where we used \eqref{eq:interpolant} in the last step. We can solve this for $\boldsymbol{\epsilon}$ and insert into the reparametrized $\mathcal{L}_{\textrm{CFM}}$ in \eqref{eq:cfm_reparam} and see that we obtain denoising score matching~\cite{vincent2011}, which implies that $\boldsymbol{\epsilon}_\theta(t,\rvx_t)$, or analogously $\rvv_\theta(t,\rvx_t)$ via their connection, learn a model of the marginal score $\nabla_{\rvx_t}\log p(\rvx_t)$.

Specifically, we have alternatively
\begin{eqnarray}
    \boldsymbol{\epsilon}_\theta(t,\rvx_t) =& -\alpha(t)\nabla_{\rvx_t}\log p(\rvx_t), \\ \label{eq:score_to_v}
    \rvv_\theta(t,\rvx_t) =& -\alpha(t)\frac{\beta(t)\dot{\alpha}(t)-\dot{\beta}(t)\alpha(t)}{\beta(t)}\nabla_{\rvx_t}\log p(\rvx_t) + \frac{\dot{\beta}(t)}{\beta(t)}\rvx_t.
\end{eqnarray}

We note that these equations only hold for a Gaussian prior without optimal transport.

\section{Related Work}
\label{sec:app-rw}

Here we discuss other approaches for unconditional molecule generation we find relevant in the context our our study that were not already discussed in Sec.~\ref{sec:rw}.
\citet{geoldm} introduces GeoLDM a geometric latent diffusion model for 3DMG. GeoLDM applies its diffusion process over a learned latent representation. So rather than updating the atom position and types in euclidean space everything is done inside the model. Similar to EDM, GeoLDM uses OpenBabel for bond prediction.
\citet{voxmol} takes a different approach than majority of prior work in representing molecules as 3D voxels rather than graphs. This is akin to 3D image processing rather than point cloud processing. This however requires a recovery process as the voxels are not a natural molecule representation. Voxels however provide a better link to the applications of vision models which majority of the diffusion framework was created for.
Lastly, \citet{GeoBFN} introduces GeoBFN a Geometric Bayesian Flow Network, that unlike diffusion models operate in the parameter space rather then product space. While the integration of 3D voxels would not work for \model, latent diffusion and BFN extensions are something relevant to future work.
\section{Megalodon Architecture}
\label{sec:arch-app}
\subsection{Architecture}
As described in ~\fref{fig:arch}, \model consists of N augmented transformer blocks that consist of a Fused Invariant Transformer (FiT) block and a structure layer.
\begin{table}[h]
    \centering
    \begin{tabular}{l|c|c}
        \hline
        \textbf{Parameter} & \textbf{Megalodon Quick} & \textbf{Megalodon} \\ \midrule
        Invariant Edge Feature Dimension & 64  & 256 \\ 
        Invariant Node Feature Dimension & 256 & 256 \\ 
        Number of Vector Features        & 64  & 128 \\ 
        Number of Layers                 & 10  & 10  \\ 
        Number of FiT Attention Heads                  & 4   & 4   \\ 
        Distance Feature Size                    & 16  & 128 \\ \hline
    \end{tabular}
    \caption{Comparison of Megalodon Quick and Megalodon hyperparameter configurations.}
    \label{tab:megalodon_comparison}
\end{table}
We refer to it as Megalodon and Megalodon Quick, as we maintain the same number of layers but weaken the representation size to achieve 2x sampling speeds compared to the base model. 

\subsubsection{Input/Output Layers}
\model takes the input molecules structures and projects them into a $N \times D$ tensor where $D$ is the number of vector features. After all augmented transformer blocks, the predicted structure is projected back down to $N \times 3$.

Similarly, the input discrete components are projected from their one hot variable to a hidden dimension size. The bonds leverage the edge feature size, and the atom types and charges use the node feature size. After all augmented transformer blocks, final prediction heads are applied to project the values back into their respective vocabulary size for discrete prediction.

\subsubsection{Fused Invariant Transformer Block}
Our Fused Invariant Transformer (FiT) block has several key differences compared to other diffusion transformers~\citep{DiT}.
\begin{itemize}
    \item Rather than just operating over the discrete atom type features $H$, we operate over a fused feature $m = \frac{1}{N} \sum_{i,j\in N} f\left( h_{\text{norm}, i, j}, h_{\text{norm}, i, j}, \text{e}_{\text{norm}, i, j}, \text{distance}_{i,j} \right) $ where $h_{\text{norm}}$ and $e_{\text{norm}}$ are the outputs of the time conditioned adaptive layer norm for the atom type and edge type features. The distance features are the concatenation of scalar distances and dot products. We note that this fusing step is important to ground the simple equivariant structure update layer to the transformer trunk.
    \item We employ query key normalization~\citep{qk_norm, esm3}.
    \item The multi-head attention is applied to $m$ to produce $\text{mha\_out}$ and then used directly in the standard feed-forward to produce $H_{out}$. To create $E_{out}$ we mimic the same steps but use $f(\text{mha\_out}_i + \text{mha\_out}_j)$ for all edges between nodes $i$ and $j$. Our feed-forward is the standard SWiGLU layer with a feature projection of 4. We note that this feed-forward for edge features is the most expensive component of the model, which is why \modelns-quick is designed the way it is. 
    
\end{itemize}

\subsubsection{Structure Layer}
Following ~\citet{diffsbdd}, the structure layer of \model consists of a single EGNN layer with a positional and cross-product update component. Before this operation, all inputs are normalized to prevent value and gradient explosion, a common problem faced when using EGNNs~\citep{egnn}. The invariant features use standard layer norm, whereas the equivariant features use an E3Norm~\citep{MiDi}.
\begin{multline}
    \label{equ:se3gnn_coord}
    \bm{x}_{i}^{l+1} = \bm{x}_{i}^{l} + \sum_{j\neq i} \frac{\bm{x}_{i}^{l} - \bm{x}_{j}^{l}}{d_{ij} + 1} \phi_x^d(\bm{h}^l_i, \bm{h}^l_j, d_{ij}^2, a_{ij}) + \\ \frac{(\bm{x}_{i}^{l} - \bar{\bm{x}}^l) \times (\bm{x}_{j}^{l} - \bar{\bm{x}}^l)}{||(\bm{x}_{i}^{l} - \bar{\bm{x}}^l) \times (\bm{x}_{j}^{l} - \bar{\bm{x}}^l)|| + 1} \phi_x^\times(\bm{h}^l_i, \bm{h}^l_j, d_{ij}^2, a_{ij}),
\end{multline}

\subsection{Compute and Data Requirements}
Similar to ~\citet{eqgat}, we use MiDi's adaptive dataloader for GEOM DRUGS with a batch cost of 200. We note that the adaptive logic randomly selects one molecule and fills in the batch with similar-sized molecules, tossing any molecules selected that do not fit the adaptive criteria out of the current epoch's available molecules. As a result, an epoch in this setting does not hold the standard connotation as time for the model to see each training data point. We use this dataloader as it was used by prior methods and we felt it important to standardize the data to best create a fair comparison. \modelns-quick is trained on 4 NVIDIA A100 GPUs for 250 epochs. \model was trained on 8 A100 GPUs for 250 epochs, taking roughly 2 days.

\modelns-flow was trained using the data splits and adaptive data loader from ~\citet{semla}, which does not discard molecules though was prefiltered to only include molecules with $\leq$ 72 atoms. It was trained for 200 epochs on 8 A100 NVIDIA GPUs.

\begin{figure}[!ht]
\centering
    \includegraphics[width=0.99\textwidth]{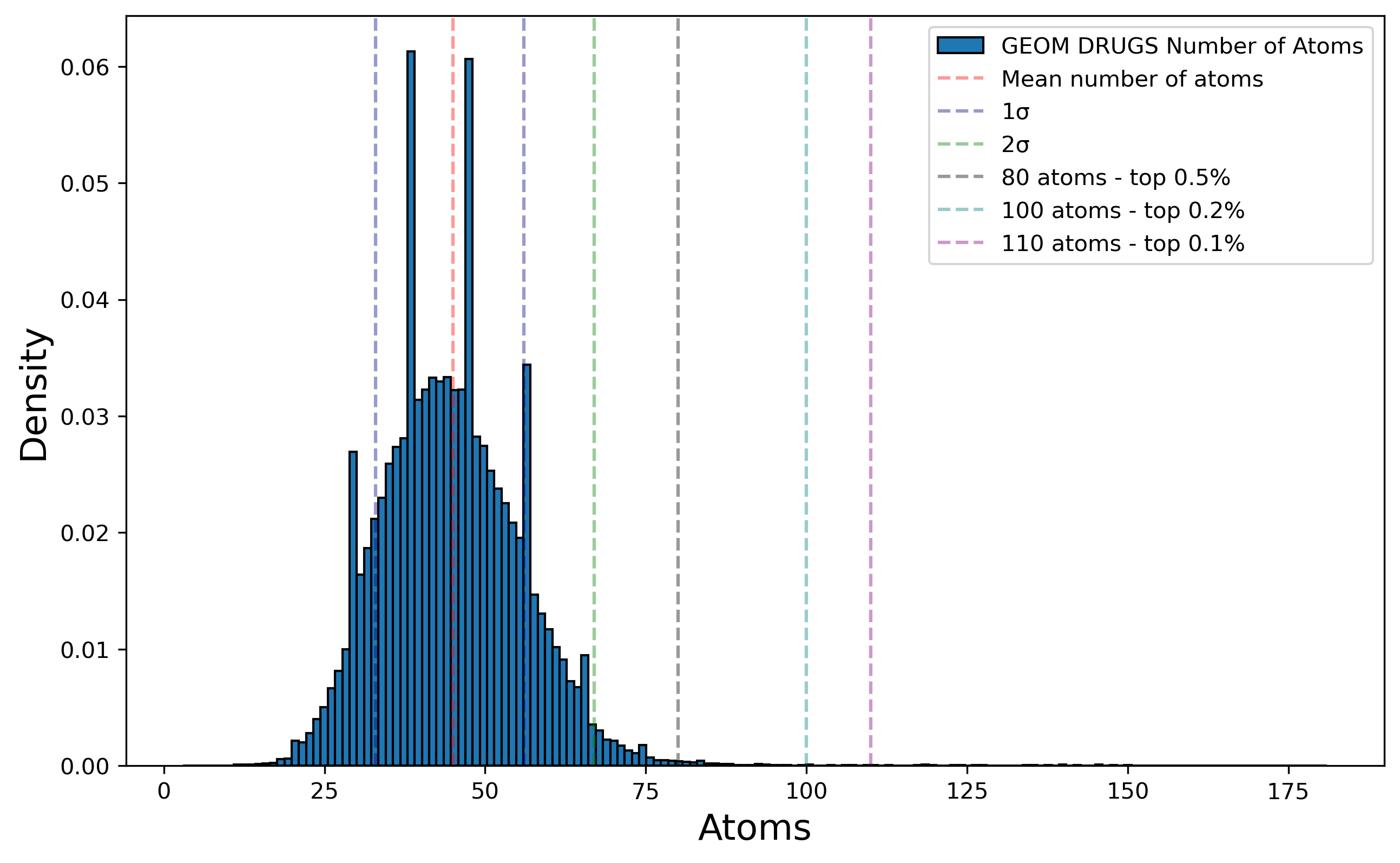}
    \caption{Distribution of molecule sizes}
    \label{fig:node-dist}
    \vspace{-2ex}
\end{figure}
\section{Extended Unconditional Generation}
\subsection{Performance on QM9}
\label{sec:qm9}

\begin{table*}[!h]
\caption{Measuring Unconditional Molecule Generation: 2D and 3D benchmarks on QM9 dataset. * Denotes taken from MiDi.}
\label{tab:qm9}
\centering
\begin{tabular}{l c | c | c c c | c c}
\toprule
& \multicolumn{1}{l|}{} & \multicolumn{1}{c|}{} & \multicolumn{3}{c|}{2D Topological ($\uparrow$)} & \multicolumn{2}{c}{3D Distributional ($\downarrow$)}  \\
Model &  & Steps & Atom Stab. & Mol Stab. & Validity  & Bond Angle & Dihedral \\ \midrule
\multicolumn{2}{l|}{MiDi*} & 500 & 0.998 & 0.975 & 0.979  & 0.670 & --  \\
\multicolumn{2}{l|}{EQGAT-diff$^{x0}_{disc}$} & 500 & 0.998 & 0.977 & 0.979 & 0.365 & 0.815 \\
\multicolumn{2}{l|}{\modelns-quick} & 500& \textbf{0.999} & \textbf{0.986} & \textbf{0.988}  & \textbf{0.241} & 0.662 \\
\multicolumn{2}{l|}{\modelns} & 500& \textbf{0.999} & \textbf{0.986} & 0.987  & 0.422 & \textbf{0.637} \\
\midrule
\multicolumn{2}{l|}{SemlaFlow} & 100 & \textbf{0.999} & \textbf{0.986} & 0.986 & 0.775 & 1.194 \\
\multicolumn{2}{l|}{\modelns-flow} & 100 & 0.998 & 0.973 & 0.976  & 0.804 & 0.970\\
\bottomrule
\end{tabular}
\vspace{-1ex}
\end{table*}

There are three popular datasets of 3D molecular structures commonly used to benchmark generative models: \textbf{QM9}, \textbf{PubChem3D}, and \textbf{GEOM Drugs}. In this work, we primarily focus on GEOM Drugs because PubChem3D provides relatively low-quality 3D structures that do not necessarily reflect low-energy conformations. Nevertheless, QM9 remains a well-established and frequently used small-scale benchmark, despite the fact that its median molecular size is unrealistically small (approximately 20 atoms). For completeness, we therefore report results on QM9 as shown in Table~\ref{tab:qm9}.

We trained SemlaFlow and EQGAT-Diff from scratch while using MiDi results from the original paper. In their original codebases, both MiDi and EQGAT-Diff were trained on molecular representations with three bond types: single, double, and triple, whereas Semla included aromatic bonds. We found that the inclusion of aromatic bonds negatively impacted molecular stability metrics, even though any molecule can be represented without them in a molecular graph using the Kekulized form. To ensure comparability, we trained all models using only single, double, and triple bonds.

In Table~\ref{tab:qm9} we observe that \modelns-quick achieves the best overall performance on QM9 in terms of both 2D (topological) and 3D (distributional) metrics. This outcome makes intuitive sense because \modelns-quick is a more lightweight variant, and smaller models often suit datasets of reduced scale, such as QM9, more effectively. In contrast, \modelns-flow appears to be too large for this dataset, leading to slightly weaker performance; however, we include it here for completeness and consistency with results on GEOM Drugs.

While flow matching benefits from fewer integration steps (here 100 steps), the diffusion-based objective with more denoising steps (in our case, 500) ultimately achieves stronger 3D quality metrics. Thus, the trade-off between computational efficiency (fewer steps) and generative fidelity (more steps) is highlighted once again in this smaller-scale setting.

\subsection{Unconditional Ablations}
\begin{table}[ht]
\caption{Measuring Unconditional Molecule Generation: 2D topological and 3D distributional benchmarks. * Denotes taken from EQGAT-Diff.}
\label{tab:3dmg-app}
\centering
\scalebox{0.78}{
\begin{tabular}{l c | c | c c c | c c}
\toprule
& \multicolumn{1}{l|}{} & \multicolumn{1}{c|}{} & \multicolumn{3}{c|}{2D Topological ($\uparrow$)} & \multicolumn{2}{c}{3D Distributional ($\downarrow$)}  \\
Model &  & Steps & Atom Stab. & Mol Stab. & Connected Validity  & Bond Angle & Dihedral \\ \midrule
\multicolumn{2}{l|}{EDM + OpenBabel*} & 1000 & 0.978 & 0.403 & 0.363  & -- & -- \\
\multicolumn{2}{l|}{MolDiff taken from \citet{MolDiff}} & 1000 & -- & -- & 0.739  & -- & -- \\
\multicolumn{2}{l|}{GeoBFN taken from \citet{GeoBFN}} & 2000 & 0.862 & 0.917 & --  & -- & -- \\
\multicolumn{2}{l|}{MiDi*} & 500 & 0.997 & 0.897 & 0.705  & -- & --  \\
\multicolumn{2}{l|}{EQGAT-diff$^{x0}_{disc}$} & 100 & 0.996 & 0.891 & 0.768 & 1.772 & 3.514 \\
\multicolumn{2}{l|}{EQGAT-diff$^{x0}_{disc}$} & 500 & 0.998 & 0.935 & 0.830 & 0.858 & 2.860 \\

\multicolumn{2}{l|}{EGNN + cross product} & 500 & 0.982 & 0.713 & 0.223  & 14.778 & 17.003 \\
\multicolumn{2}{l|}{\modelns-quick} & 500& 0.998 & 0.961 & 0.900  & 0.689 & 2.383 \\
\multicolumn{2}{l|}{\modelns} & 100& 0.998 & 0.939 & 0.817  & 0.871 & 3.367 \\
\multicolumn{2}{l|}{\modelns} & 500& \textbf{0.999} & 0.977 & 0.927  & \textbf{0.461} & \textbf{1.231} \\
\midrule
\multicolumn{2}{l|}{SemlaFlow} & 20 & 0.997 & 0.962 & 0.875 & 2.188 & 3.173 \\
\multicolumn{2}{l|}{SemlaFlow} & 100 & 0.998 & 0.979 & 0.920 & 1.274 & 1.934 \\
\multicolumn{2}{l|}{\modelns-flow} & 20 & 0.996 & 0.964 & 0.886  & 1.892 & 3.184 \\
\multicolumn{2}{l|}{\modelns-flow} & 100 & 0.997 & \textbf{0.990} & \textbf{0.948}  & 0.976 & 2.085 \\
\bottomrule
\end{tabular}
}
\end{table}

We include each primary model in its base form as well as with 5x fewer inference steps. The flow models do not have to be retrained as they were trained to learn a continuous vector field, whereas the diffusion models must be retrained due to the change in variance discretization in the forward diffusion process.

We also include EGNN + cross product which is similar to \model except the transformer layers were replaced by the standard invariant and edge feature updates in~\citet{egnn}.
Prior methods exist that improve upon EDM + Open Babel and maintain that bonds are generated external to the model via Open Babel~\citep{EquiFM}. We do not include such methods in our comparison as, for the most part, public code with weights is not available, and Open Babel introduces significant bias and errors, which make evaluating the model difficult~\citep{EquiFM, pat_walters_blog}.

Open Babel, while a powerful tool for molecular manipulation and conversion, can introduce several potential errors, particularly in the context of bond assignment and 3D structure generation. Some common errors include:
\begin{itemize}
    \item Incorrect bond orders: Open Babel often assigns bond orders based on geometric heuristics or atom types, which can lead to inaccuracies, especially in complex or exotic molecules where bond orders are not trivial.
    \item Geometric distortions: When converting between different formats or generating 3D coordinates, Open Babel may generate suboptimal or distorted geometries, especially if the input structure is incomplete or poorly defined.
    \item Protonation state assumptions: Open Babel may incorrectly infer or standardize protonation states, which can lead to chemical inaccuracies, especially in sensitive systems such as drug-like molecules or biologically active compounds.
    \item Ambiguous aromaticity: Open Babel can sometimes misinterpret or incorrectly assign aromaticity, which can lead to an incorrect representation of the molecular structure.
    \item Missing stereochemistry: While converting or generating structures, stereochemistry can be incorrectly assigned or lost altogether, affecting the overall molecular properties.
\end{itemize}

\subsection{3D Distributional Metrics}

To evaluate the geometric fidelity of the generated molecules, we compute the Wasserstein-1 distance between the generated and target distributions of bond angles, following the methodology of \citep{eqgat}. The overall bond angle metric is defined as:

\begin{equation*}
W_{\text{angles}} = \sum_{y \in \text{atom types}} p(y) \cdot W_1\big( \hat{D}_{\text{angle}}(y), D_{\text{angle}}(y) \big),
\end{equation*}

where $ p(y) $ is the probability of atom type $ y $, $ W_1 $ denotes the Wasserstein-1 distance, $ \hat{D}_{\text{angle}}(y) $ is the bond angle distribution for atom type $ y $ in the generated data, and $ D_{\text{angle}}(y) $ is the corresponding distribution in of test set.

Similarly, for torsion angles, the metric is calculated as:

\begin{equation*}
W_{\text{torsions}} = \sum_{y \in \text{bond types}} p(y) \cdot W_1\big( \hat{D}_{\text{torsion}}(y), D_{\text{torsion}}(y) \big),
\end{equation*}

where $ p(y) $ is the probability of bond type $ y $, $ \hat{D}_{\text{torsion}}(y) $ is the torsion angle distribution for bond type $ y $ in the generated data, and $ D_{\text{torsion}}(y) $ is the corresponding distribution in the test set. Since we utilized RDKit to identify torsions, the torsional distribution difference was computed only for valid molecules.

\section{Geometric Conformational Analysis Benchmarks}
\label{sec:energy-app}


To quantitatively evaluate the fidelity of generated molecular structures relative to their relaxed counterparts, we introduce benchmarks that assess differences in bond lengths, bond angles, and torsion angles. These metrics provide detailed insights into how closely the generated conformations approximate local minima on the potential energy surface.

\subsection{Bond Length Differences}

For each bond in the molecule, we compute the difference in bond lengths between the initial (generated) and optimized (relaxed) structures. Let $ r_{ij}^{\text{init}}$ and $r_{ij}^{\text{opt}}$ denote the distances between atoms $i$ and $j$ in the initial and optimized conformations, respectively. The bond length difference $\Delta r_{ij}$ is calculated as:

\begin{equation*}
    \Delta r_{ij} = \left| r_{ij}^{\text{init}} - r_{ij}^{\text{opt}} \right|
\end{equation*}

We compute average differences and corresponding frequencies for each possible combination of source atom type, bond type, and target atom type. The final result is the weighted sum of those differences. 

\subsection{Bond Angle Differences}

For each bond angle formed by three connected atoms $i$, $j$, and $k$, we calculate the angle difference between the initial and optimized structures. Let $ \theta_{ijk}^{\text{init}}$ and $ \theta_{ijk}^{\text{opt}} $ represent the bond angles at atom $j$ in the initial and optimized conformations, respectively. The bond angle difference $\Delta \theta_{ijk}$ is given by:

\begin{equation*}
\Delta \theta_{ijk} = \min\left( \left| \theta_{ijk}^{\text{init}} - \theta_{ijk}^{\text{opt}} \right|, 180^\circ - \left| \theta_{ijk}^{\text{init}} - \theta_{ijk}^{\text{opt}} \right|\right)
\end{equation*}

As with bond lengths, these differences are grouped based on the types of atoms and bonds involved to calculate the final results.

\subsection{Torsion Angle Differences}

Torsion angles involve four connected atoms $i$, $j$,  $k$, and $l$. We compute the difference in torsion angles between the initial and optimized structures using:

\begin{equation*}
\Delta \phi_{ijkl} = \min\left( \left| \phi_{ijkl}^{\text{init}} - \phi_{ijkl}^{\text{opt}} \right|,\ 360^\circ - \left| \phi_{ijkl}^{\text{init}} - \phi_{ijkl}^{\text{opt}} \right| \right)
\end{equation*}

where $\phi_{ijkl}^{\text{init}}$ and $\phi_{ijkl}^{\text{opt}}$ are the dihedral angles in the initial and optimized conformations, respectively. This formula accounts for the periodicity of dihedral angles, ensuring the smallest possible difference is used.

By analyzing these statistical measures, we can assess the structural deviations of generated molecules from their relaxed forms. Lower average differences indicate that the generative model produces conformations closer to local energy minima.

As with bond lengths, these differences are grouped based on the types of atoms and bonds involved to calculate the final results.

\subsection{xTB Energy benchmark}

We also computed the median and mean relaxation energies ($ \Delta E_{\text{relax}} $) for both ground truth data and generated molecules using both GFN2-xTB and MMFF force fields. The relaxation energy is defined as the energy difference between the optimized (relaxed) structure and the initial (generated) structure:

\begin{equation*}
\Delta E_{\text{relax}} = E_{\text{optimized}} - E_{\text{initial}}
\end{equation*}

\subsubsection{Limitations of MMFF for Evaluating GFN2-xTB Structures}

Previous studies have used the MMFF force field to assess the quality of generated molecular structures~\citep{geodiff, semla}. However, the choice of force field is critical when evaluating molecular geometries, as different force fields can yield significantly different energy landscapes. For ground truth conformers optimized using GFN2-xTB, the mean relaxation energy difference $\Delta E_{\text{relax}}$ calculated with GFN2-xTB is nearly zero, as expected. However, when these same structures are evaluated using the MMFF force field, the mean $\Delta E_{\text{relax}}$ is approximately $16\,\text{kcal/mol}$, which aligns with literature values reporting MMFF errors in the range of $15$–$20\,\text{kcal/mol}$ when compared to higher-level methods like GFN2-xTB~\citep{foloppe2019energy}. In other words, xTB is significantly more accurate than MMFF, especially for data generated with xTB.

In contrast, our generated molecules exhibit a mean $\Delta E_{\text{relax}}$ of around $5\,\text{kcal/mol}$ when relaxed with GFN2-xTB, significantly smaller than the error observed when using MMFF. This demonstrates that our model produces structures that are much closer to the GFN2-xTB energy minima compared to what MMFF evaluations suggest. These substantial energy discrepancies—stemming from systematic differences like optimal bond lengths and angles—highlight that MMFF is inappropriate for evaluating structures optimized or generated within the GFN2-xTB framework. Relying on MMFF can, therefore, lead to misleading assessments of structural quality.

\subsubsection{Achieving Thermodynamically Relevant Energy Accuracy}

In statistical thermodynamics, conformers exist in dynamic equilibrium, and their population distribution is determined by their relative free energies. The equilibrium constant $K$ between two conformers is given by:

\begin{equation*}
K = e^{- \Delta G^\circ / RT},
\end{equation*}

where $\Delta G^\circ$ is the standard free energy difference, $R$ is the gas constant, and $T$ is the temperature in Kelvin. At room temperature ($298~\text{K}$), the thermal energy $RT$ is approximately $0.6~\text{kcal/mol}$. A free energy difference of $1.36~\text{kcal/mol}$ corresponds to a tenfold difference in the equilibrium constant. The GEOM dataset selects conformers within a $2.5~\text{kcal/mol}$ energy window, encompassing about $99.9\%$ of the Boltzmann population for the lowest-energy conformers.

Our generated molecules have a median relaxation energy $\Delta E_{\text{relax}}$ of around $3 \text{kcal/mol}$, approaching this thermally relevant interval. Notably, our model is the first to achieve such proximity to the thermodynamic threshold, establishing a significant milestone in the generative modeling of molecular conformations. By generating conformations with relaxation energies close to the thermal energy interval, our model effectively produces energetically feasible structures near local minima on the GFN2-xTB potential energy surface. This breakthrough demonstrates the model's potential for practical applications, such as conformational searches and drug discovery, where accurate conformer generation is crucial.

By measuring these geometric deviations and appropriate relaxation energy, we offer a comprehensive evaluation of the accuracy of generated molecular conformations, facilitating the development of more precise generative models in computational chemistry.

\begin{table}[!ht]
\caption{xTB Relaxation Error: Length \AA, angles degrees, energy kcal/mol. These metrics are taken over the valid molecules from~\tref{tab:3dmg}. Methods are grouped by model type: diffusion (500 steps) and flow matching (100 steps)}
\label{tab:xtb-app}
\centering
\scalebox{0.78}{
\begin{tabular}{l c | c c c | c c c c}
\toprule
Model &  & Bond Length & Bond Angles & Dihedral & Median $\Delta E_{\text{relax}}$ & Mean $\Delta E_{\text{relax}}$ & Mean $\Delta E_{\text{relax}}^{\text{MMFF}}$ \\ \midrule
\multicolumn{2}{l|}{GEOM-Drugs} & 0.0 & 0.0 & 7.2e-3 & 0.00 & 1.0e-3 & 16.48\\
\midrule
\multicolumn{2}{l|}{EQGAT-diff} & 0.0076 & 0.95 & 7.98 & 6.36 & 11.06 & 28.45 \\

\multicolumn{2}{l|}{\modelns-small} & 0.0085 & 0.88 & 7.28 & 5.78 & 9.74 & 24.87 \\
\multicolumn{2}{l|}{\modelns} & \textbf{0.0061} & \textbf{0.66} & \textbf{5.42} & \textbf{3.17} & \textbf{5.71} & 21.61  \\
\midrule
\multicolumn{2}{l|}{SemlaFlow} & 0.0309 & 2.03 & 6.01 & 32.96 & 93.13 & 69.46 \\
\multicolumn{2}{l|}{\modelns-flow} & \textbf{0.0225} & \textbf{1.59} & \textbf{5.49} & \textbf{20.86} & \textbf{46.86}  & 45.51\\
\bottomrule
\end{tabular}
}
\end{table}

\section{Megalodon Molecule Visualization}

\begin{figure}[H]
    \centering
    \includegraphics[width=1\linewidth, angle=0]{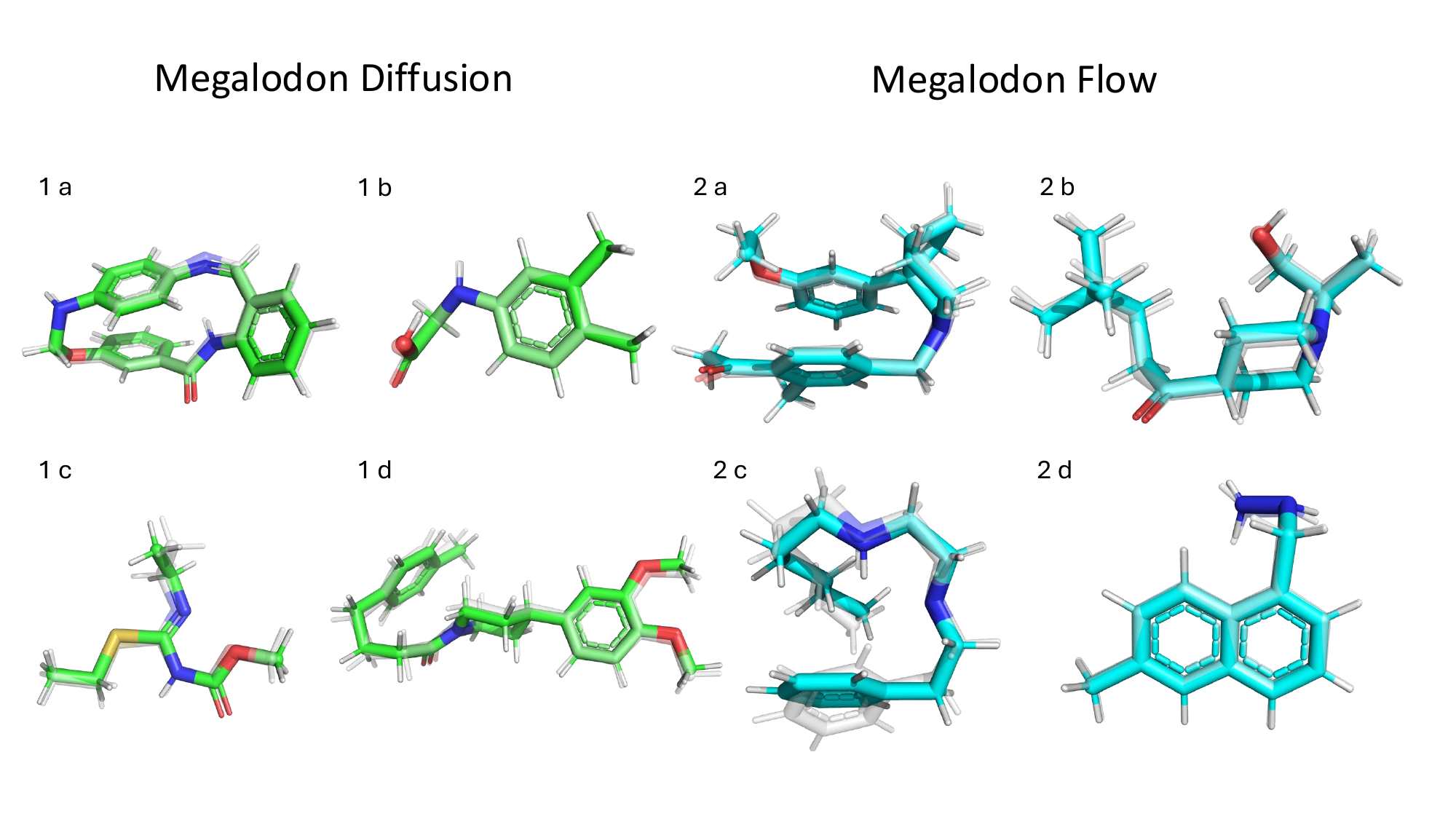}
    \caption{Examples of generated molecules using Megalodon: (1) Diffusion and (2) Flow Matching. Each generated molecule is displayed alongside its corresponding optimized structure (shown in transparent grey). The examples include small aromatic molecules (1b, 2d), molecules exhibiting pi-stacking interactions (1a, 2a), non-aromatic molecules (1c, 2b), and a molecule with a macrocycle (1a).}
    \label{fig:generated_molecules}
\end{figure}
\section{Limitations}

While we show that \model performs well across a variety of 3D de novo molecules tasks there are still some limitations that are worthy of discussion.

\begin{itemize}
    \item \model like ~\citet{eqgat} and the prior edge prediction generative models before it relies on maintaining $N^2$ edge features, which is quite expensive. Recently~\citep{semla} was able to avoid this issue for a majority of the model architecture by fusing the edge and atom features, but this creates a trade-off between model speed and accuracy. Our ablations show that the larger edge features are critical for strong energy performance, so it is still an open question for how to best deal with discrete edge types as each atom can have a maximum of 6 bonds at a time, so is needing to model all $N$ potential pairings at all times really necessary? We leave future work to explore this in greater depth.
    \item As discussed herein, the existing 3D molecule generation benchmarks are quite limited. A common theme that has been discussed in prior work~\citep{eqgat, semla}. While we make strides in expanding the field of view of de novo design and energy-based benchmarks. More work needs to be done to measure important qualities, as even for common conditional design benchmarks, metrics such as QED are not meaningful in practice, and even more complex properties like protein-ligand binding affinity can be directly optimized for with non-3D structure-based methods~\citep{evosbdd}. For these reasons, we looked to explore conditional structure generation, but across the board, small molecule benchmarking is a current field-wide limitation when compared to the current drug discovery practices. 
    \item A general limitation for all prior 3DMG models including \model is that they cannot generalize to unseen atom types due to the one hot representation during training. As a result, these models can only be used for GEOM-Drug-like molecules. While fairly extensive, it is a limitation worth noting as the flexibility is limited when compared to 2D or SMILES-based LLMs.
\end{itemize}

\end{document}